\documentclass{article}

\usepackage{microtype}
\usepackage{graphicx}
\usepackage{chngcntr}
\usepackage{subcaption}
\usepackage{booktabs} %
\usepackage{bm}
\usepackage{xcolor}
\usepackage{listings}

\usepackage{hyperref}

\usepackage{stmaryrd}

\usepackage[accepted]{icml2020}
\usepackage{booktabs}       %
\usepackage{amsfonts}       %
\usepackage{microtype}      %
\usepackage{url}
\usepackage{verbatim} %
\usepackage{graphicx}
\usepackage{caption} 
\usepackage{multirow}
\usepackage{xspace}
\usepackage{epsfig}
\usepackage{amsmath}
\usepackage{amsthm}
\usepackage{amssymb}
\usepackage{times}
\usepackage{xr}
\usepackage{bbm}
\usepackage{dsfont}
\usepackage{bm}
\usepackage{enumitem}
\usepackage{hyperref}       %

\newcommand{\multiline}[3][11mm]{\begin{minipage}[c][#1]{#2}#3\end{minipage}}

\newcommand{\xhdr}[1]{{\noindent\bfseries #1}.}

\newcommand{\cut}[1]{}

\newcommand\figref[1]{Figure~\ref{#1}}
\newcommand\tabref[1]{Table~\ref{#1}}

\newcommand{\norm}[1]{\lVert#1\rVert}

\newcommand\domain[1]{\textsc{#1}}

\newcommand{\simulator}{s}
\newcommand{\dynamics}{d}
\newcommand{\encoder}{\textsc{Encoder}}
\newcommand{\processor}{\textsc{Processor}}
\newcommand{\decoder}{\textsc{Decoder}}

\newenvironment{itemize*}%
  {\begin{itemize}[topsep=0pt, partopsep=0pt]%
    \setlength{\itemsep}{0pt}%
    \setlength{\parskip}{0pt}}%
  {\end{itemize}}
  
\newenvironment{enumerate*}%
  {\begin{enumerate}[topsep=-5pt, partopsep=0pt]%
    \setlength{\itemsep}{0pt}%
    \setlength{\parskip}{0pt}%
    \setlength{\parsep}{0pt}}%
  {\end{enumerate}}

\newcommand\viddambreak[1]{\href{http://tny.sh/QpFGhIK}{#1}}
\newcommand\vidmulti[1]{\href{http://tny.sh/bf5FJnN}{#1}}
\newcommand\vidcont[1]{\href{http://tny.sh/lLvCAc5}{#1}}
\newcommand\vidobs[1]{\href{http://tny.sh/MfmIF9H}{#1}}
\newcommand\vidshake[1]{\href{http://tny.sh/dGoSxSA}{#1}}
\newcommand\vidboxbath[1]{\href{http://tny.sh/mrOkfpg}{#1}}
\newcommand\vidgenvortex[1]{\href{http://tny.sh/tWNrMBg}{#1}}
\newcommand\vidgenramps[1]{\href{http://tny.sh/2fmmUfN}{#1}}
\newcommand\vidgenmulti[1]{\href{http://tny.sh/C0TUQH7}{#1}}
\newcommand\vidcconv[1]{\href{http://tny.sh/6I0girh}{#1}}
\newcommand\vidthreedsand[1]{\href{http://tny.sh/4WbbnXU}{#1}}
\newcommand\vidthreedgoop[1]{\href{http://tny.sh/Ojm7Olb}{#1}}
\newcommand\vidthreedwater[1]{\href{http://tny.sh/jFszr4x}{#1}}
\newcommand\vidtwodsand[1]{\href{http://tny.sh/2M64scu}{#1}}
\newcommand\vidtwodgoop[1]{\href{http://tny.sh/uImt11M}{#1}}
\newcommand\vidtwodwater[1]{\href{http://tny.sh/xlDxA0A}{#1}}
\newcommand\vidfail[1]{\href{http://tny.sh/Tc01u0R}{#1}}
\newcommand\vidthreedwaters[1]{\href{http://tny.sh/water3ds}{#1}}
\newcommand\vidsandramps[1]{\href{http://tny.sh/sandramps}{#1}}
\newcommand\vidwaterramps[1]{\href{http://tny.sh/waterramps}{#1}}

\definecolor{codegreen}{rgb}{0,0.4,0}
\definecolor{codegray}{rgb}{0.4,0.4,0.4}
\definecolor{codepurple}{rgb}{0.5,0,0.7}
\definecolor{backcolour}{rgb}{0.96,0.96,0.94}
 
\lstdefinestyle{mystyle}{
    backgroundcolor=\color{backcolour},   
    commentstyle=\color{codegreen},
    keywordstyle=\color{magenta},
    numberstyle=\tiny\color{codegray},
    stringstyle=\color{codepurple},
    basicstyle=\ttfamily\footnotesize,
    breakatwhitespace=false,         
    breaklines=true,                 
    captionpos=b,                    
    keepspaces=true,                 
    numbers=left,                    
    numbersep=5pt,                  
    showspaces=false,                
    showstringspaces=false,
    showtabs=false,                  
    tabsize=2
}
 
\def \papertitle{Learning to Simulate Complex Physics with Graph Networks}
\icmltitlerunning{Learning to Simulate}
\begin{document}

\twocolumn[

\icmltitle{\papertitle}

\icmlsetsymbol{equal}{*}

\begin{icmlauthorlist}
\icmlauthor{Alvaro Sanchez-Gonzalez}{equal,deepmind}
\icmlauthor{Jonathan Godwin}{equal,deepmind}
\icmlauthor{Tobias Pfaff}{equal,deepmind}
\icmlauthor{Rex Ying}{equal,deepmind,stanford}
\icmlauthor{Jure Leskovec}{stanford}
\icmlauthor{Peter W. Battaglia}{deepmind}
\end{icmlauthorlist}

\icmlaffiliation{stanford}{Department of Computer Science, Stanford University, Stanford, CA, USA}
\icmlaffiliation{deepmind}{DeepMind, London, UK}

\icmlcorrespondingauthor{}{{alvarosg@google.com}}
\icmlcorrespondingauthor{}{{jonathangodwin@google.com}}
\icmlcorrespondingauthor{}{{tpfaff@google.com}}
\icmlcorrespondingauthor{}{{rexying@stanford.edu}}
\icmlcorrespondingauthor{}{{jure@cs.stanford.edu}}
\icmlcorrespondingauthor{}{{peterbattaglia@google.com}}

\icmlkeywords{Machine Learning, Graph Neural Networks, Physical Simulation, Particle-based Fluid Simulation}

\vskip 0.3in
]

\printAffiliationsAndNotice{\icmlEqualContribution} %

\begin{abstract}
Here we present a machine learning framework and model implementation that can learn to simulate a wide variety of challenging physical domains, involving fluids, rigid solids, and deformable materials interacting with one another.
Our framework---which we term ``Graph Network-based Simulators'' (GNS)---represents the state of a physical system with particles, expressed as nodes in a graph, and computes dynamics via learned message-passing. 
Our results show that our model can generalize from single-timestep predictions with thousands of particles during training, to different initial conditions, thousands of timesteps, and at least an order of magnitude more particles at test time. 
Our model was robust to hyperparameter choices across various evaluation metrics: the main determinants of long-term performance were the number of message-passing steps, and mitigating the accumulation of error by corrupting the training data with noise. 
Our GNS framework advances the state-of-the-art in learned physical simulation, and holds promise for solving a wide range of complex forward and inverse problems.
\end{abstract}

\begin{figure}[t]
  \centering
  \includegraphics[width=0.48\textwidth]{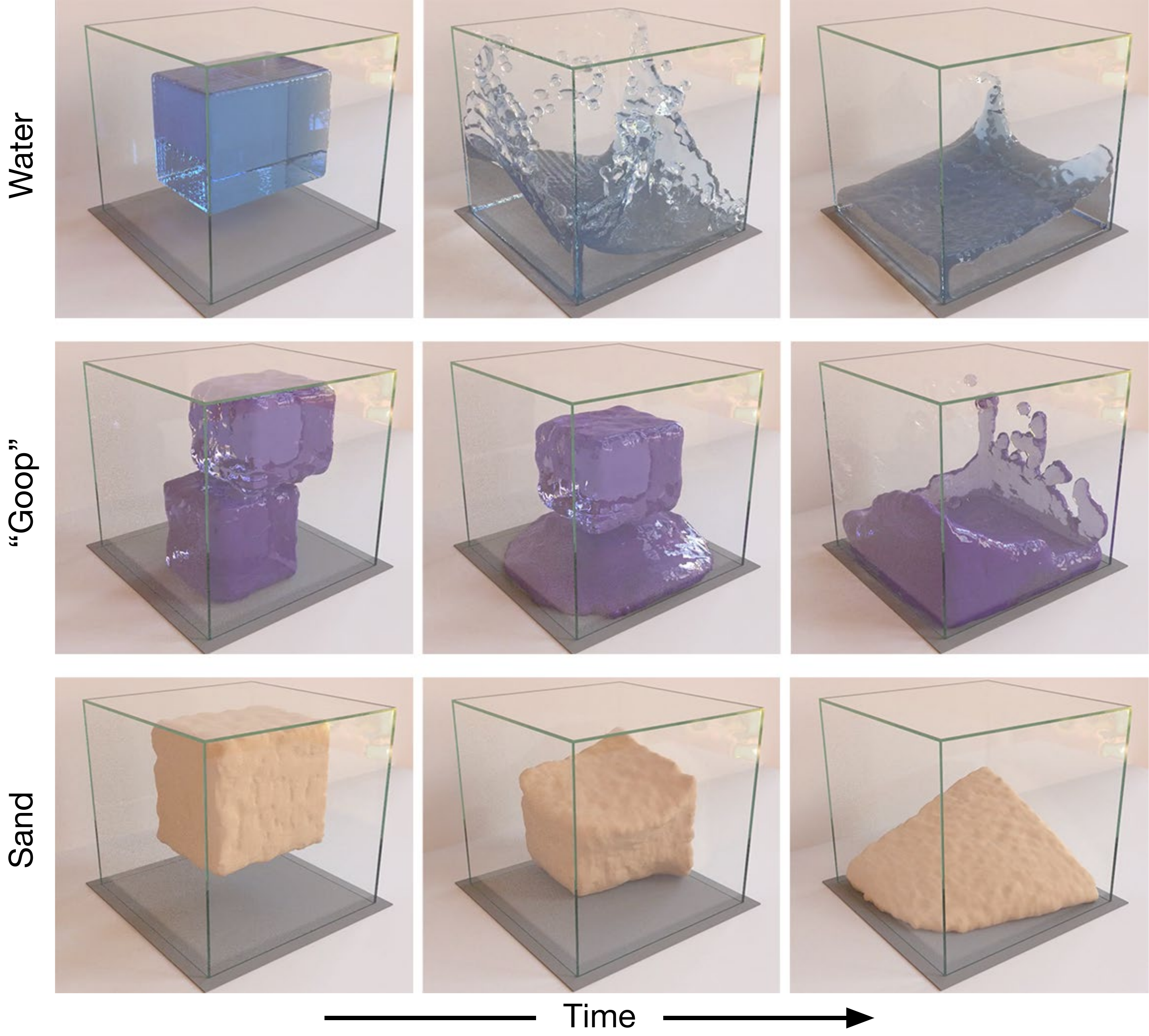}
\caption{Rollouts of our GNS model for our \domain{Water-3D}, \domain{Goop-3D} and \domain{Sand-3D} datasets. It learns to simulate rich materials at resolutions sufficient for high-quality rendering \vidthreedwater{[video]}.} 
  \label{fig:teaser}
\end{figure}

\begin{figure*}[t]
  \centering
  \includegraphics[height=\textwidth,angle=90,origin=c, trim=210mm 0 40mm 0]{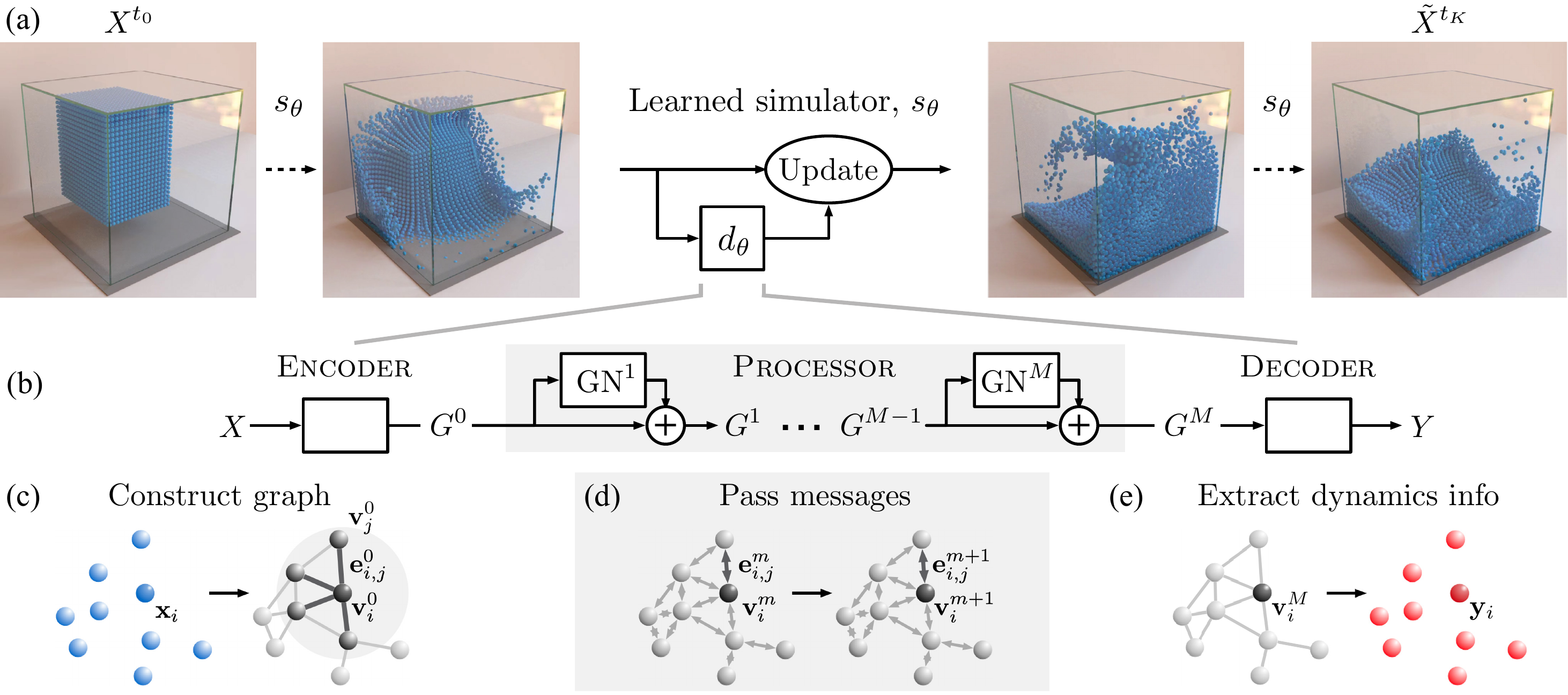}
  \caption{
  \textbf{(a)} Our GNS predicts future states represented as particles using its learned dynamics model, $d_\theta$, and a fixed update procedure.
  \textbf{(b)} The $d_\theta$ uses an ``encode-process-decode'' scheme, which computes dynamics information, $Y$, from input state, $X$.
  \textbf{(c)} The \encoder{} constructs latent graph, $G^0$, from the input state, $X$. 
  \textbf{(d)} The \processor{} performs $M$ rounds of learned message-passing over the latent graphs, $G^0, \dots, G^M$. 
  \textbf{(e)} The \decoder{} extracts dynamics information, $Y$, from the final latent graph, $G^M$.}
  \label{fig:schematic}
\end{figure*}

\section{Introduction}

Realistic simulators of complex physics are invaluable to many scientific and engineering disciplines, however traditional simulators can be very expensive to create and use. Building a simulator can entail years of engineering effort, and often must trade off generality for accuracy in a narrow range of settings. High-quality simulators require substantial computational resources, which makes scaling up prohibitive. 
Even the best are often inaccurate due to insufficient knowledge of, or difficulty in approximating, the underlying physics and parameters.
An attractive alternative to traditional simulators is to use machine learning to train simulators directly from observed data, however the large state spaces and complex dynamics have been difficult for standard end-to-end learning approaches to overcome.

Here we present a powerful machine learning framework for learning to simulate complex systems from data---``Graph Network-based Simulators'' (GNS). Our framework imposes strong inductive biases, where rich physical states are represented by graphs of interacting particles, and complex dynamics are approximated by learned message-passing among nodes.

We implemented our GNS framework in a single deep learning architecture, and found it could learn to accurately simulate a wide range of physical systems in which fluids, rigid solids, and deformable materials interact with one another. Our model also generalized well to much larger systems and longer time scales than those on which it was trained. While previous learning simulation approaches~\cite{li2018learning,ummenhofer2020lagrangian} have been highly specialized for particular tasks, we found our single GNS model performed well across dozens of experiments and was generally robust to hyperparameter choices. Our analyses showed that performance was determined by a handful of key factors: its ability to compute long-range interactions, inductive biases for spatial invariance, and training procedures which mitigate the accumulation of error over long simulated trajectories.
\section{Related Work}
\label{sec:related}

Our approach focuses on \textit{particle-based} simulation, which is used widely across science and engineering, e.g., computational fluid dynamics, computer graphics. States are represented as a set of particles, which encode mass, material, movement, etc. within local regions of space. Dynamics are computed on the basis of particles' interactions within their local neighborhoods. One popular particle-based method for simulating fluids is ``smoothed particle hydrodynamics'' (SPH)~\cite{monaghan1992smoothed}, which evaluates pressure and viscosity forces %
around each particle, and updates particles' velocities and positions accordingly. Other techniques, such as ``position-based dynamics'' (PBD)~\cite{muller2007position} and ``material point method'' (MPM)~\cite{sulsky1995application}, are more suitable for interacting, deformable materials. In PBD, incompressibility and collision dynamics involve resolving pairwise distance constraints between particles, and directly predicting their position changes. 
Several differentiable particle-based simulators have recently appeared, e.g., DiffTaichi~\cite{hu2019difftaichi}, PhiFlow~\cite{holl2020learning}, and Jax-MD~\cite{schoenholz2019jax}, which can backpropagate gradients through the architecture.

Learning simulations from data~\cite{grzeszczuk1998neuroanimator} has been an important area of study with applications in physics and graphics. Compared to engineered simulators, a learned simulator can be far more efficient for predicting complex phenomena~\cite{he2019learning}; e.g., \cite{ladicky2015data,wiewel2019latent} learn parts of a fluid simulator for faster prediction.

Graph Networks (GN)~\cite{battaglia2018relational}---a type of graph neural network~\cite{scarselli2008graph}---have recently proven effective at learning forward dynamics in various settings that involve interactions between many entities. A GN maps an input graph to an output graph with the same structure but potentially different node, edge, and graph-level attributes, and can be trained to learn a form of message-passing~\cite{gilmer2017neural}, where latent information is propagated between nodes via the edges. GNs and their variants, e.g., ``interaction networks'', can learn to simulate rigid body, mass-spring, n-body, and robotic control systems~\cite{battaglia2016interaction,chang2016compositional,sanchez2018graph,mrowca2018flexible,li2019propagation,sanchez2019hamiltonian}, as well as non-physical systems, such as multi-agent dynamics \cite{tacchetti2018relational,sun2019stochastic}, algorithm execution \cite{velickovic2020Neural}, and other dynamic graph settings~\cite{trivedi2018dyrep,trivedi2017know,yan2018spatial,manessi2020dynamic}. 

Our GNS framework builds on and generalizes several lines of work, especially~\citet{sanchez2018graph}'s GN-based model which was applied to various robotic control systems,~\citet{li2018learning}'s DPI which was applied to fluid dynamics, and~\citet{ummenhofer2020lagrangian}'s Continuous Convolution (CConv) which was presented as a non-graph-based method for simulating fluids. Crucially, our GNS framework is a \textit{general} approach to learning simulation, is simpler to implement, and is more accurate across fluid, rigid, and deformable material systems.

\section{GNS Model Framework}
\label{sec:model}

\subsection{General Learnable Simulation}
\label{sec:model:general}

We assume $X^t \in \mathcal{X}$ is the state of the world at time $t$. Applying  physical dynamics over $K$ timesteps yields a trajectory of states, $\mathbf{X}^{t_{0:K}} = (X^{t_0}, \dots, X^{t_K})$.
A \textit{simulator}, ${\simulator: \mathcal{X} \rightarrow \mathcal{X}}$, models the dynamics by mapping preceding states to causally consequent future states. We denote a simulated ``rollout'' trajectory as, ${\tilde{\mathbf{X}}^{t_{0:K}} = (X^{t_0}, \tilde{X}^{t_1}, \dots, \tilde{X}^{t_K})}$, which is computed iteratively by, ${\tilde{X}^{t_{k+1}} = \simulator(\tilde{X}^{t_{k}})}$
for each timestep.
Simulators compute dynamics information that reflects how the current state is changing, and use it to update the current state to a predicted future state (see Figure~\ref{fig:schematic}(a)). An example is a numerical differential equation solver: the equations compute dynamics information, i.e., time derivatives, and the integrator is the update mechanism.

A learnable simulator, $\simulator_\theta$, computes the dynamics information with a parameterized function approximator, ${\dynamics_\theta: \mathcal{X} \rightarrow \mathcal{Y}}$, whose parameters, $\theta$, can be optimized for some training objective. The ${Y \in \mathcal{Y}}$ represents the dynamics information, whose semantics are determined by the update mechanism. The update mechanism can be seen as a function which takes the $\tilde{X}^{t_{k}}$, and uses $d_\theta$ to predict the next state, $\tilde{X}^{t_{k+1}} = \textrm{Update}(\tilde{X}^{t_{k}}, \dynamics_\theta)$. Here we assume a simple update mechanism---an Euler integrator---and $\mathcal{Y}$ that represents accelerations. However, more sophisticated update procedures which call $\dynamics_\theta$ more than once can also be used, such as higher-order integrators (e.g.,~\citet{sanchez2019hamiltonian}). 

\subsection{Simulation as Message-Passing on a Graph}
\label{sec:model:message-passing}

Our learnable simulation approach adopts a particle-based representation of the physical system (see Section~\ref{sec:related}), i.e., $X = (\mathbf{x}_0, \dots, \mathbf{x}_N)$, where each of the $N$ particles' $\mathbf{x}_i$ represents its state. Physical dynamics are approximated by interactions among the particles, e.g., exchanging energy and momentum among their neighbors. The way particle-particle interactions are modeled determines the quality and generality of a simulation method---i.e., the types of effects and materials it can simulate, in which scenarios the method performs well or poorly, etc. We are interested in learning these interactions, which should, in principle, allow learning the dynamics of any system that can be expressed as particle dynamics. So it is crucial that different $\theta$ values allow $d_\theta$ to span a wide range of particle-particle interaction functions.

Particle-based simulation can be viewed as message-passing on a graph. The nodes correspond to particles, and the edges correspond to pairwise relations among particles, over which interactions are computed. 
We can understand methods like SPH in this framework---the messages passed between nodes could correspond to, e.g., evaluating pressure using the density kernel.

We capitalize on the correspondence between particle-based simulators and message-passing on graphs to define a general-purpose $d_\theta$ based on GNs. Our $d_\theta$ has three steps---\encoder, \processor, \decoder~\cite{battaglia2018relational} (see Figure~\ref{fig:schematic}(b)).

\xhdr{\encoder{} definition} The ${\encoder: \mathcal{X} \rightarrow \mathcal{G}}$ embeds the particle-based state representation, $X$, as a latent graph, ${G^0 = \encoder(X)}$, where $G = (V, E, \mathbf{u})$, $\mathbf{v}_i \in V$, and $\mathbf{e}_{i,j} \in E$ (see Figure~\ref{fig:schematic}(b,c)). 
The node embeddings, $\mathbf{v}_i = \varepsilon^v(\mathbf{x}_i)$, are learned functions of the particles' states.
Directed edges are added to create paths between particle nodes which have some potential interaction. The edge embeddings, $\mathbf{e}_{i,j} = \varepsilon^e(\mathbf{r}_{i,j})$, are learned functions of the pairwise properties of the corresponding particles, $\mathbf{r}_{i,j}$, e.g., displacement between their positions, spring constant, etc.
The graph-level embedding, $\mathbf{u}$, can represent global properties such as gravity and magnetic fields (though in our implementation we simply appended those as input node features---see Section~\ref{sec:experiment:architecture} below).

\xhdr{\processor{} definition} The ${\processor: \mathcal{G} \rightarrow \mathcal{G}}$ computes interactions among nodes via $M$ steps of learned message-passing, to generate a sequence of updated latent graphs, ${\mathbf{G} = (G^1, ..., G^M)}$, where ${G^{m+1} = \mathrm{GN}^{m+1}(G^m)}$ (see Figure~\ref{fig:schematic}(b,d)). It returns the final graph, ${G^M = \processor(G^0)}$. Message-passing allows information to propagate and constraints to be respected: the number of message-passing steps required will likely scale with the complexity of the interactions.

\xhdr{\decoder{} definition} The ${\decoder: \mathcal{G} \rightarrow \mathcal{Y}}$ extracts dynamics information from the nodes of the final latent graph, ${\mathbf{y}_i = \delta^v(\mathbf{v}^M_i)}$ (see Figure~\ref{fig:schematic}(b,e)).
Learning $\delta^v$ should cause the $\mathcal{Y}$ representations to reflect relevant dynamics information, such as acceleration, in order to be semantically meaningful to the update procedure.

\section{Experimental Methods}
\label{sec:experiments}

Code and data available at \href{https://github.com/deepmind/deepmind-research/tree/master/learning_to_simulate}{github.com/deepmind/deepmind-research/tree/master/learning\_to\_simulate}.

\subsection{Physical Domains}
\label{sec:experiment:domains}

We explored how our GNS learns to simulate in datasets which contained three diverse, complex physical materials: water as a barely damped fluid, chaotic in nature; sand as a granular material with complex frictional behavior; and ``goop'' as a viscous, plastically deformable material. 
These materials have very different behavior, and in most simulators, require implementing separate material models or even entirely different simulation algorithms.

For one domain, we use~\citet{li2018learning}'s \domain{BoxBath}, which simulates a container of water and a cube floating inside, all represented as particles, using the PBD engine FleX~\cite{macklin2014unified}. 

We also created \domain{Water-3D}, a high-resolution 3D water scenario with randomized water position, initial velocity and volume, comparable to~\citet{ummenhofer2020lagrangian}'s containers of water. We used SPlisHSPlasH~\cite{bender2015dfsph}, a SPH-based fluid simulator with strict volume preservation to generate this dataset.

For most of our domains, we use the Taichi-MPM engine~\cite{hu2018mpm} to simulate a variety of challenging 2D and 3D scenarios. We chose MPM for the simulator because it can simulate a very wide range of materials, and also has some different properties than PBD and SPH, e.g., particles may become compressed over time.

Our datasets typically contained 1000 train, 100 validation and 100 test trajectories, each simulated for 300-2000 timesteps (tailored to the average duration for the various materials to come to a stable equilibrium). A detailed listing of all our datasets can be found in the Supplementary Materials~\ref{supp:sec:experiments}.

\subsection{GNS Implementation Details}
\label{sec:experiment:architecture}

We implemented the components of the GNS framework using standard deep learning building blocks, and used standard nearest neighbor algorithms \cite{dong2011efficient,chen2009fast,tang2016visualizing} to construct the graph.

\xhdr{Input and output representations}
Each particle's input state vector represents position, a sequence of $C=5$ previous velocities\footnote{$C$ is a hyperparameter which we explore in our experiments.}, and features that capture static material properties (e.g., water, sand, goop, rigid, boundary particle), $\mathbf{x}^{t_k}_i = [\mathbf{p}^{t_k}_i, \dot{\mathbf{p}}^{t_{k-C+1}}_i, \dots, \dot{\mathbf{p}}^{t_k}_i, \mathbf{f}_i]$, respectively. 
The global properties of the system, $\mathbf{g}$, include external forces and global material properties, when applicable.
The prediction targets for supervised learning are the per-particle average acceleration, $\ddot{\mathbf{p}}_i$.
Note that in our datasets, we only require $\mathbf{p}_i$ vectors: the $\dot{\mathbf{p}}_i$ and $\ddot{\mathbf{p}}_i$ are computed from $\mathbf{p}_i$ using finite differences. For full details of these input and target features, see Supplementary Material Section~\ref{supp:sec:experiments}.

\xhdr{\encoder{} details} The \encoder{} constructs the graph structure $G^0$ by assigning a node to each particle and adding edges between particles within a ``connectivity radius'', $R$, which reflected local interactions of particles, and which was kept constant for all simulations of the same resolution. For generating rollouts, on each timestep the graph's edges were recomputed by a nearest neighbor algorithm, to reflect the current particle positions.

The \encoder{} implements $\varepsilon^v$ and $\varepsilon^e$ as multilayer perceptrons (MLP), which encode node features and edge features into the latent vectors, $\mathbf{v}_i$ and $\mathbf{e}_{i,j}$, of size 128.

We tested two \encoder{} variants, distinguished by whether they use absolute versus relative positional information. For the absolute variant, the input to $\varepsilon^v$ was the $\mathbf{x}_i$ described above, with the globals features concatenated to it. The input to $\varepsilon^e$, i.e., $\mathbf{r}_{i,j}$, did not actually carry any information and was discarded, with the $\mathbf{e}^0_i$ in $G^0$ set to a trainable fixed bias vector. The relative \encoder{} variant was designed to impose an inductive bias of invariance to absolute spatial location. The $\varepsilon^v$ was forced to ignore $\mathbf{p}_i$ information within $\mathbf{x}_i$ by masking it out. The $\varepsilon^e$ was provided with the relative positional displacement, and its magnitude\footnote{Similarly, relative velocities could be used to enforce invariance to inertial frames of reference.}, ${\mathbf{r}_{i,j} = [(\mathbf{p}_i - \mathbf{p}_j), \norm{\mathbf{p}_i - \mathbf{p}_j}]}$. Both variants concatenated the global properties $\mathbf{g}$ onto each $\mathbf{x}_i$ before passing it to $\varepsilon^v$.

\xhdr{\processor{} details} Our processor uses a stack of $M$ GNs (where $M$ is a hyperparameter) with identical structure, MLPs as internal edge and node update functions, and either shared or unshared parameters (as analyzed in Results Section~\ref{sec:results:ablations}). We use GNs without global features or global updates (similar to an interaction network)\footnote{In preliminary experiments we also attempted using a \processor{} with a full GN and a global latent state, for which the global features $\mathbf{g}$ are encoded with a separate $\varepsilon^g$ MLP.}, and with a residual connections between the input and output latent node and edge attributes.

\xhdr{\decoder{} details} Our decoder's learned function, $\delta^v$, is an MLP. After the \decoder{}, the future position and velocity are updated using an Euler integrator, so the $\mathbf{y}_i$ corresponds to accelerations, $\ddot{\mathbf{p}}_i$, with 2D or 3D dimension, depending on the physical domain. As mentioned above, the supervised training outputs were simply these, $\ddot{\mathbf{p}}_i$ vectors\footnote{Note that in this case optimizing for acceleration is equivalent to optimizing for position, because the acceleration is computed as first order finite difference from the position and we use an Euler integrator to update the position.}.

\xhdr{Neural network parameterizations} All MLPs have two hidden layers (with ReLU activations), followed by a non-activated output layer, each layer with size of 128. All MLPs (except the output decoder) are followed by a LayerNorm~\cite{ba2016layer} layer, which we generally found improved training stability.

\subsection{Training}
\label{sec:experiment:training}

\xhdr{Software} We implemented our models using TensorFlow 1, Sonnet 1, and the ``Graph Nets'' library~(\citeyear{graphnetlib}).

\xhdr{Training noise}
Modeling a complex and chaotic simulation system requires the model to mitigate error accumulation over long rollouts. Because we train our models on ground-truth one-step data, they are never presented with input data %
corrupted by this sort of accumulated noise.
This means that when we generate a rollout by feeding the model with its own noisy, previous predictions as input, the fact that its inputs are outside the training distribution may lead it to make more substantial errors, and thus rapidly accumulate further error.
We use a simple approach to make the model more robust to noisy inputs: at training we corrupt the input velocities of the model with random-walk noise $\mathcal{N}(0, \sigma_v=0.0003)$ (adjusting input positions), so the training distribution is closer to the distribution generated during rollouts. See Supplementary Materials~\ref{supp:sec:experiments} for full details.

\xhdr{Normalization}
We normalize all input and target vectors elementwise to zero mean and unit variance, using statistics computed online during training.
Preliminary experiments showed that normalization led to faster training, though converged performance was not noticeably improved.

\xhdr{Loss function and optimization procedures}
We randomly sampled particle state pairs $(\mathbf{x}^{t_k}_i, \mathbf{x}^{t_{k+1}}_i)$ from training trajectories, calculated target accelerations $\ddot{\mathbf{p}}^{t_{k}}_i$ (subtracting the noise added to the most recent input velocity), and computed the $L_2$ loss on the predicted per-particle accelerations, i.e., $L(\mathbf{x}^{t_k}_i, \mathbf{x}^{t_{k+1}}_i; \theta) = \norm{ d_{\theta}(\mathbf{x}^{t_{k}}_i) - \ddot{\mathbf{p}}^{t_{k}}_i}^2$.
We optimized the model parameters $\theta$ over this loss with the Adam optimizer~\cite{kingma2014adam}, using a nominal\footnote{The actual batch size varies at each step dynamically. See Supplementary Material for more details.} mini-batch size of 2. We performed a maximum of 20M gradient update steps, with exponential learning rate decay from $10^{-4}$ to $10^{-6}$.
While models can train in significantly less steps, we avoid aggressive learning rates to reduce variance across datasets and make comparisons across settings more fair.

We evaluated our models regularly during training by producing full-length rollouts on 5 held-out validation trajectories, and recorded the associated model parameters for best rollout MSE. We stopped training when we observed negligible decrease in MSE, which, on GPU/TPU hardware, was typically within a few hours for smaller, simpler datasets, and up to a week for the larger, more complex datasets.

\subsection{Evaluation}
\label{sec:experiment:evaluation}

To report quantitative results, we evaluated our models after training converged by computing one-step and rollout metrics on held-out test trajectories, drawn from the same distribution of initial conditions used for training.
We used particle-wise MSE as our main metric between ground truth and predicted data, both for rollout and one-step predictions, averaging across time, particle and spatial axes. We also investigated distributional metrics including optimal transport (OT) \cite{villani2003topics} (approximated by the Sinkhorn Algorithm \cite{cuturi2013sinkhorn}), and Maximum Mean Discrepancy (MMD) \cite{gretton2012kernel}. 
For the generalization experiments we also evaluate our models on a number of initial conditions drawn from distributions different than those seen during training, including, different number of particles, different object shapes, different number of objects, different initial  positions and velocities and longer trajectories.
See Supplementary Materials~\ref{supp:sec:experiments} for full details on metrics and evaluation.

\begin{table}[t!]
\begin{small}
\begin{center}
\begin{tabular}[p]{|l||c|c|c|c|}
	\hline
	\multiline{20mm}{\textbf{Experimental\\domain}} & 
	\multiline{4mm}{\centering$N$} & 
	\multiline{4mm}{\centering$K$}&
	\multiline{10mm}{\centering\textbf{1-step} \\ ($\times10^{-9}$)}  &
	\multiline{10mm}{\centering\textbf{Rollout} \\ ($\times10^{-3}$)}\\\hline\hline
	\vidthreedwater{\domain{Water-3D}} (SPH) & 13k & 800 & 8.66 & 10.1 \\\hline
    \vidthreedsand{\domain{Sand-3D}}  & 20k & 350 & 1.42 & 0.554 \\\hline
    \vidthreedgoop{\domain{Goop-3D}}  & 14k & 300 & 1.32 & 0.618 \\\hline
    \vidthreedwaters{\domain{Water-3D-S}} (SPH) & 5.8k & 800 & 9.66 & 9.52 \\\hline
    \vidboxbath{\domain{BoxBath}} (PBD) & 1k & 150 & 54.5 & 4.2 \\\hline\hline
    \vidtwodwater{\domain{Water}}  & 1.9k & 1000 & 2.82 & 17.4 \\\hline
    \vidtwodsand{\domain{Sand}}  & 2k & 320 & 6.23 & 2.37 \\\hline
    \vidtwodgoop{\domain{Goop}}  & 1.9k & 400 & 2.91 & 1.89 \\\hline
    \vidmulti{\domain{MultiMaterial}}  & 2k & 1000 & 1.81 & 16.9 \\\hline
    \vidshake{\domain{FluidShake}}  & 1.3k & 2000 & 2.1 & 20.1 \\\hline
    \viddambreak{\domain{WaterDrop}}  & 1k & 1000 & 1.52 & 7.01 \\\hline
    \viddambreak{\domain{WaterDrop-XL}}  & 7.1k & 1000 & 1.23 & 14.9 \\\hline
    \vidwaterramps{\domain{WaterRamps}}  & 2.3k & 600 & 4.91 & 11.6 \\\hline
    \vidsandramps{\domain{SandRamps}}  & 3.3k & 400 & 2.77 & 2.07 \\\hline
    \vidobs{\domain{RandomFloor}}  & 3.4k & 600 & 2.77 & 6.72 \\\hline
    \vidcont{\domain{Continuous}}  & 4.3k & 400 & 2.06 & 1.06 \\\hline
\end{tabular}
\end{center}
\end{small}
\caption{
\label{fig:mse}List of maximum number of particles $N$, sequence length $K$, and 
quantitative model accuracy (MSE) on the held-out test set. All domain names are also hyperlinks to the \href{https://sites.google.com/view/learning-to-simulate}{video website}.}
\end{table} 
\section{Results}
\label{sec:results}

\begin{figure*}[t]
  \centering
  \includegraphics[width=\textwidth]{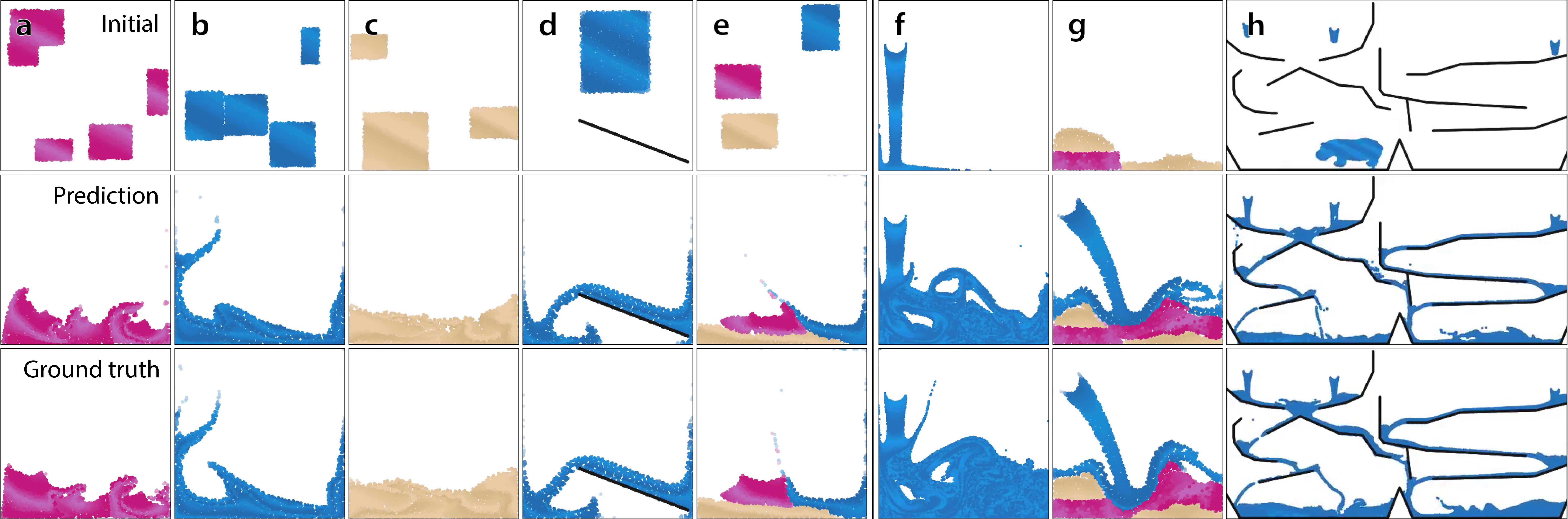}
  \caption{
  We can simulate many materials, from \textbf{(a)} \domain{Goop} over \textbf{(b)} \domain{Water} to \textbf{(c)} \domain{Sand}, and \textbf{(d)} their interaction with rigid obstacles (\domain{WaterRamps}). We can even train a single model on \textbf{(e)} multiple materials and their interaction (\domain{MultiMaterial}). We applied pre-trained models on several out-of-distribution tasks, involving \textbf{(f)} high-res turbulence (trained on \domain{WaterRamps}), \textbf{(g)} multi-material interactions with unseen objects (trained on \domain{MultiMaterial}), and \textbf{(h)} generalizing on significantly larger domains (trained on \domain{WaterRamps}). In the two bottom rows, we show a comparison of our model's prediction with the ground truth on the final frame for goop and sand, and on a representative mid-trajectory frame for water.}
  \label{fig:materials}
\end{figure*}

Our main findings are that our GNS model can learn accurate, high-resolution, long-term simulations of different fluids, deformables, and rigid solids, and it can generalize well beyond training to much longer, larger, and challenging settings. In Section~\ref{sec:results:baselines} below, we compare our GNS model to two recent, related approaches, and find our approach was simpler, more generally applicable, and more accurate.

To challenge the robustness of our architecture, we used a single set of model hyperparameters for training across all of our experiments.
Our GNS architecture used the relative \encoder{} variant, 10 steps of message-passing, with unshared GN parameters in the \processor{}.
We applied noise with a scale of $3\cdot 10^{-4}$ to the input states during training.

\subsection{Simulating Complex Materials}
\label{sec:results:single}

Our GNS model was very effective at learning to simulate different complex materials.
Table~\ref{fig:mse} shows the one-step and rollout accuracy, as MSE, for all experiments. For intuition about what these numbers mean, the edge length of the container was approximately $1.0$, and \figref{fig:materials}(a-c)
shows rendered images of the rollouts of our model, compared to ground truth\footnote{All rollout videos can be found here: \url{https://sites.google.com/view/learning-to-simulate}}. Visually, the model's rollouts are quite plausible. Though specific model-generated trajectories can be distinguished from ground truth when compared side-by-side, it is difficult to visually classify individual videos as generated from our model versus the ground truth simulator. 

Our GNS model scales to large amounts of particles and very long rollouts. With up to 19k particles in our 3D domains---substantially greater than demonstrated in previous methods---GNS can operate at resolutions high enough for practical prediction tasks and high-quality 3D renderings (e.g., \figref{fig:teaser}). And although our models were trained to make one-step predictions, the long-term trajectories remain plausible even over thousands of rollout timesteps. %

The GNS model could also learn how the materials respond to unpredictable external forces. In the \domain{FluidShake} domain, a container filled with water is being moved side-to-side, causing splashes and irregular waves. 

Our model could also simulate fluid interacting with complicated static obstacles, as demonstrated by our \domain{WaterRamps} and \domain{SandRamps} domains in which water or sand pour over 1-5 obstacles. \figref{fig:materials}(d) depicts comparisons between our model and ground truth, and \tabref{fig:mse} shows quantitative performance measures. 

We also trained our model on continuously varying material parameters. In the \domain{Continuous} domain, we varied the friction angle of a granular material, to yield behavior similar to a liquid (0$^{\circ}$), sand (45$^{\circ}$), or gravel ($>$ 60$^{\circ}$). Our results and \vidcont{videos} show that our model can account for these continuous variations, and even interpolate between them: a model trained with the region $[30^{\circ}, 55^{\circ}]$ held out in training can accurately predict within that range. Additional quantitative results are available in Supplementary Materials~\ref{supp:sec:results}.

\subsection{Multiple Interacting Materials}
\label{sec:results:multiple}

So far we have reported results of training identical GNS architectures separately on different systems and materials. However, we found we could go a step further and train a single architecture with a single set of parameters to simulate all of our different materials, interacting with each other in a single system.

In our \domain{MultiMaterial} domain, the different materials could interact with each other in complex ways, which means the model had to effectively learn the product space of different interactions (e.g., water-water, sand-sand, water-sand, etc.). The behavior of these systems was often much richer than the single-material domains: the stiffer materials, such as sand and goop, could form temporary semi-rigid obstacles, which the water would then flow around. \figref{fig:materials}(e) and \vidmulti{this video} shows renderings of such rollouts.
Visually, our model's performance in \domain{MultiMaterial} is comparable to its performance when trained on those materials individually.

\begin{figure*}[ht]
  \begin{tabular}[p]{l|r}
  \begin{minipage}{4.23in}
  \includegraphics[trim=6.5mm 0 0 0,width=\textwidth]{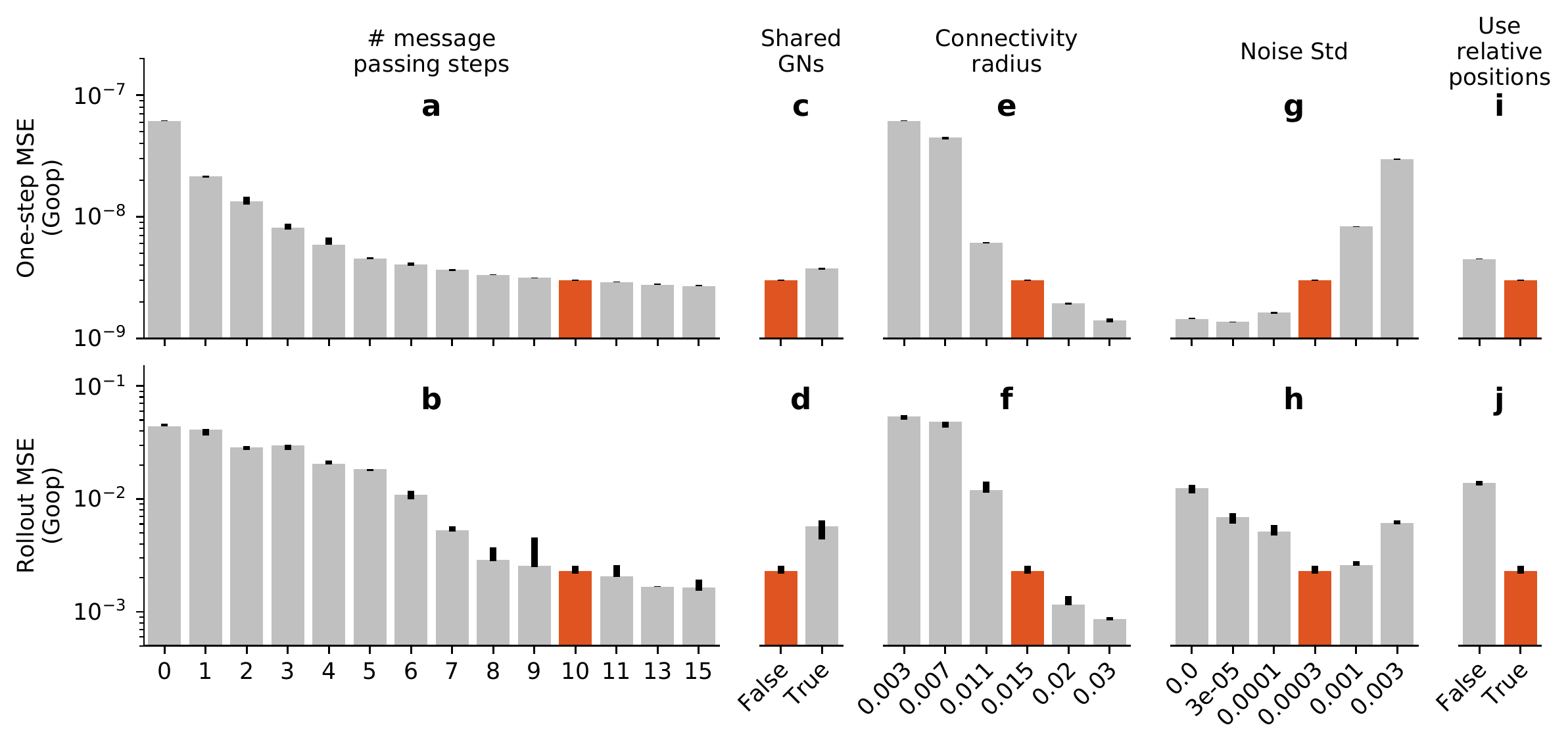}%
  \end{minipage}
& {\begin{minipage}{2.3in}
  \includegraphics[width=1.\textwidth]{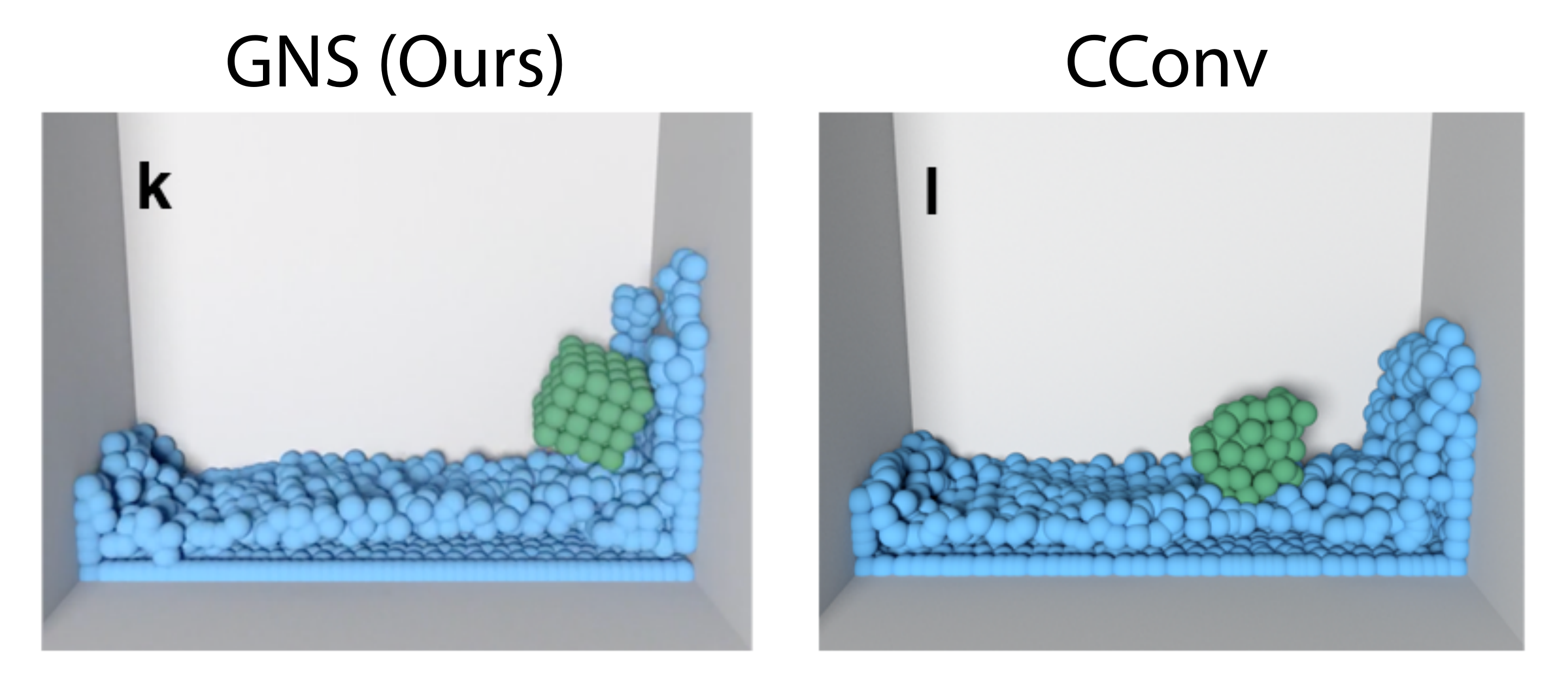} \\ 
  \vspace{1mm}
  \includegraphics[trim=0 0 0 0, width=1.0\textwidth]{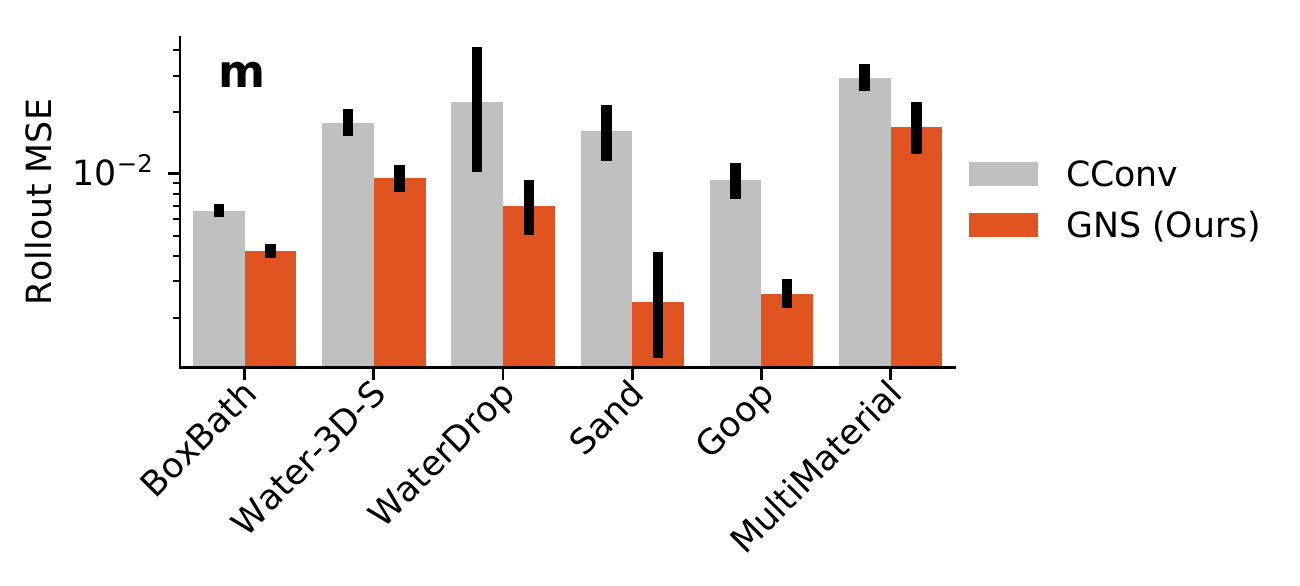}
  \end{minipage}}
\end{tabular}

  \caption{(left) Effect of different ablations (grey) against our model (red) on the one-step error \textbf{(a,c,e,g,i)} and the rollout error \textbf{(b,d,f,h,j)}. Bars show the median seed performance averaged across the entire \domain{Goop} test dataset. Error bars display lower and higher quartiles, and are shown for the default parameters. (right) Comparison of average performance of our GNS model to CConv. \textbf{(k,l)} Qualitative comparison between GNS \textbf{(k)} and CConv \textbf{(l)} in \domain{BoxBath} after 50 rollout steps (\vidcconv{video link}). \textbf{(m)} Quantitative comparison of our GNS model (red) to the CConv model (grey) across the test set . For our model, we trained one or more seeds using the same set of hyper-parameters and show results for all seeds. For the CConv model we ran several variations including different radius sizes, noise levels, and number of unroll steps during training, and show the result for the best seed. Errors bars show the standard error of the mean across all of the trajectories in the test set (95\% confidence level).}
  \label{fig:main_ablations}%
\end{figure*}

\subsection{Generalization}
\label{sec:results:generalization}
We found that the GNS generalizes well even beyond its training distributions, which suggests it learns a more general-purpose understanding of the materials and physical processes experienced during training.

To examine its capacity for generalization, we trained a GNS architecture on \domain{WaterRamps}, whose initial conditions involved a square region of water in a container, with 1-5 ramps of random orientation and location. After training, we tested the model on several very different settings. In one generalization condition, rather than all water being present in the initial timestep, we created an ``inflow'' that continuously added water particles to the scene during the rollout, as shown in \figref{fig:materials}(f). When unrolled for 2500 time steps, the scene contained 28k particles---an order of magnitude more than the 2.5k particles used in training---and the model was able to predict complex, highly chaotic dynamics not experienced during training, as can be seen in \vidgenvortex{this video}. The predicted dynamics were visually similar to the ground truth sequence.

Because we used relative displacements between particles as input to our model, in principle the model should handle scenes with much larger spatial extent at test time. We evaluated this on a much larger domain, with several inflows over a complicated arrangement of slides and ramps (see \figref{fig:materials}(h), video \vidgenramps{here}). The test domain's spatial width $\times$ height were $8.0 \times 4.0$, which was 32x larger than the training domain's area; at the end of the rollout, the number of particles was 85k, which was 34x more than during training; we unrolled the model for 5000 steps, which was 8x longer than the training trajectories. We conducted a similar experiment with sand on the \domain{SandRamps} domain, testing model generalization to hourglass-shaped ramps. 

As a final, extreme test of generalization, we applied a model trained on \domain{MultiMaterial} to a custom test domain with inflows of various materials and shapes (\figref{fig:materials}(g)). The model learned about frictional behavior between different materials (sand on sticky goop, versus slippery floor), and that the model generalized well to unseen shapes, such as hippo-shaped chunks of goop and water, falling from mid-air, as can be observed in this \vidgenmulti{video}.

\subsection{Key Architectural Choices}
\label{sec:results:ablations}

We performed a comprehensive analysis of our GNS's architectural choices to discover what influenced performance most heavily. We analyzed a number of hyperparameter choices---e.g., number of MLP layers, linear encoder and decoder functions, global latent state in the \processor{}---but found these had minimal impact on performance (see Supplementary Materials~\ref{supp:sec:results} for details).

While our GNS model was generally robust to architectural and hyperparameter settings, we also identified several factors which had more substantial impact: 
\begin{enumerate*}
    \item the number of message-passing steps,
    \item shared vs. unshared \processor{} GN parameters,
    \item the connectivity radius, 
    \item the scale of noise added to the inputs during training,
    \item relative vs. absolute \encoder{}.
\end{enumerate*}
We varied these choices systematically for each axis, fixing all other axes with the default architecture's choices, and report their impact on model performance in the \domain{Goop} domain (\figref{fig:main_ablations}).

For (1), \figref{fig:main_ablations}(a,b) shows that a greater number of message-passing steps $M$ yielded improved performance in both one-step and rollout accuracy. This is likely because increasing $M$ allows computing longer-range, and more complex, interactions among particles. Because computation time scales linearly with $M$, in practice it is advisable to use the smallest $M$ that still provides desired performance.

For (2), \figref{fig:main_ablations}(c,d) shows that models with unshared GN parameters in the \processor{} yield better accuracy, especially for rollouts. Shared parameters imposes a strong inductive bias that makes the \processor{} analogous to a recurrent model, while unshared parameters are more analogous to a deep architecture, which incurs $M$ times more parameters. In practice, we found marginal difference in computational costs or overfitting, so we conclude that using unshared parameters has little downside.

For (3), \figref{fig:main_ablations}(e,f) shows that greater connectivity $R$ values yield lower error. Similar to increasing $M$, larger neighborhoods allow longer-range communication among nodes. Since the number of edges increases with $R$, more computation and memory is required, so in practice the minimal $R$ that gives desired performance should be used.

For (4), we observed that rollout accuracy is best for an intermediate noise scale (see \figref{fig:main_ablations}(g,h)), consistent with our motivation for using it (see Section~\ref{sec:experiment:training}). We also note that one-step accuracy decreases with increasing noise scale. This is not surprising: adding noise makes the training distribution less similar to the uncorrupted distribution used for one-step evaluation. 

For (5), \figref{fig:main_ablations}(i,j) shows that the relative \encoder{} is clearly better than the absolute version. This is likely because the underlying physical processes that are being learned are invariant to spatial position, and the relative \encoder{}'s inductive bias is consistent with this invariance.

\subsection{Comparisons to Previous Models}
\label{sec:results:baselines}

We compared our approach to two recent papers which explored learned fluid simulators using particle-based approaches. \citet{li2018learning}'s~DPI studied four datasets of fluid, deformable, and solid simulations, and presented four different, distinct architectures, which were similar to~\citet{sanchez2018graph}'s, with additional features such as as hierarchical latent nodes. 
When training our GNS model on DPI's \domain{BoxBath} domain, we found it could learn to simulate the rigid solid box floating in water, faithfully maintaining the stiff relative displacements among rigid particles, as shown \figref{fig:main_ablations}(k) and \vidboxbath{this video}. Our GNS model did not require any modification---the box particles' material type was simply a feature in the input vector---while DPI required a specialized hierarchical mechanism and forced all box particles to preserve their relative displacements with each other. Presumably the relative \encoder{} and training noise alleviated the need for such mechanisms.

\citet{ummenhofer2020lagrangian}'s CConv propagates information across particles\footnote{The authors state CConv does not use an explicit graph representation, however we believe their particle update scheme can be interpreted as a special type of message-passing on a graph. See Supplementary Materials~\ref{supp:sec:baselines}.}, and uses particle update functions and training procedures which are carefully tailored to modeling fluid dynamics (e.g., an SPH-like local kernel, different sub-networks for fluid and boundary particles, a loss function that weights slow particles with few neighbors more heavily). 
\citet{ummenhofer2020lagrangian} reported CConv outperformed DPI, so we quantitatively compared our GNS model to CConv. 
We implemented CConv as described in its paper, plus two additional versions which borrowed our noise and multiple input states,
and performed hyperparameter sweeps over various CConv parameters. \figref{fig:main_ablations}(m) shows that across all six domains we tested, our GNS model with default hyperparameters has better rollout accuracy than the best CConv model (among the different versions and hyperparameters) for that domain. In \vidcconv{this comparison video}, we observe than CConv performs well for domains like water, which it was built for, but struggles with some of our more complex materials. Similarly, in a CConv rollout of the \domain{BoxBath domain} the rigid box loses its shape (\figref{fig:main_ablations}(l)), while our method preserves it. See Supplementary Materials~\ref{supp:sec:baselines} for full details of our DPI and CConv comparisons. 

\section{Conclusion}
\label{sec:conclusion}

We presented a powerful machine learning framework for learning to simulate complex systems, based on particle-based representations of physics and learned message-passing on graphs.
Our experimental results show our single GNS architecture can learn to simulate the dynamics of fluids, rigid solids, and deformable materials, interacting with one another, using tens of thousands of particles over thousands time steps. We find our model is simpler, more accurate, and has better generalization than previous approaches.

While here we focus on mesh-free particle methods, our GNS approach may also be applicable to data represented using meshes, such as finite-element methods. There are also natural ways to incorporate stronger, generic physical knowledge into our framework, such as Hamiltonian mechanics~\cite{sanchez2019hamiltonian} and rich, architecturally imposed symmetries. To realize advantages over traditional simulators, future work should explore how to parameterize and implement GNS computations more efficiently, and exploit the ever-improving parallel compute hardware. Learned, differentiable simulators will be valuable for solving inverse problems, by not strictly optimizing for forward prediction, but for inverse objectives as well.

More broadly, this work is a key advance toward more sophisticated generative models, and furnishes the modern AI toolkit with a greater capacity for physical reasoning. 
\section*{Acknowledgements}

We thank Victor Bapst, Jessica Hamrick and our reviewers for valuable feedback on the work and manuscript, and we thank Benjamin Ummenhofer for advice on implementing the continuous convolution baseline model.

\bibliography{refs}
\bibliographystyle{icml2020}

\cut{
\appendix
\section{Do \emph{not} have an appendix here}

\textbf{\emph{Do not put content after the references.}}
Put anything that you might normally include after the references in a separate
supplementary file.

We recommend that you build supplementary material in a separate document.
If you must create one PDF and cut it up, please be careful to use a tool that
doesn't alter the margins, and that doesn't aggressively rewrite the PDF file.
pdftk usually works fine. 

\textbf{Please do not use Apple's preview to cut off supplementary material.} In
previous years it has altered margins, and created headaches at the camera-ready
stage. 
}

\appendix

\counterwithin{figure}{section}
\counterwithin{table}{section}
\counterwithin{algorithm}{section}
\onecolumn
\icmltitle{Supplementary Material: \papertitle}

\section{Supplementary GNS Model Details}
\label{supp:sec:model}

\xhdr{Update mechanism}
Our GNS implementation here uses semi-implicit Euler integration to update the next state based on the predicted accelerations:
\begin{align*}
    \dot{\mathbf{p}}^{t_{k+1}} &=\dot{\mathbf{p}}^{t_{k}} + \Delta t\cdot\ddot{\mathbf{p}}^{t_{k}} \\
    \mathbf{p}^{t_{k+1}} &=\mathbf{p}^{t_{k}} + \Delta t\cdot\dot{\mathbf{p}}^{t_{k+1}}
\end{align*}
where we assume $\Delta t = 1$ for simplicity. We use this in contrast to forward Euler ($\mathbf{p}^{t_{k+1}} =\mathbf{p}^{t_{k}} + \Delta t\cdot\dot{\mathbf{p}}^{t_{k}}$) so the acceleration $\ddot{\mathbf{p}}^{t_{k}}$ predicted by the model can directly influence $\mathbf{p}^{t_{k+1}}$.

\xhdr{Optimizing parameters of learnable simulator}
Learning a simulator $\simulator_\theta$, can in general be expressed as optimizing its parameters $\theta$ over some objective function,
\begin{align*}
    \theta^* &\leftarrow \arg_\theta\min \mathbb{E}_{\mathbb{P}(\mathbf{X}^{t_{0:K}})} L(\mathbf{X}^{t_{1:K}}, \tilde{\mathbf{X}}^{t_{1:K}}_{\simulator_\theta, X^{t_0}})\,.
\end{align*}
${\mathbb{P}(\mathbf{X}^{t_{0:K}})}$ represents a distribution over state trajectories, starting from the initial conditions $X^{t_0}$, over $K$ timesteps.
The $\tilde{\mathbf{X}}^{t_{1:K}}_{\simulator_\theta, X^{t_0}}$ indicates the simulated rollout generated by $\simulator_\theta$ given $X^{t_0}$.
The objective function $L$, considers the whole trajectory generated by $\simulator_\theta$.
In this work, we specifically train our GNS model on a one-step loss function, $L_{\text{1-step}}$, with
\begin{align*}
\theta^*_{\text{1-step}} &\leftarrow \arg_\theta\min \mathbb{E}_{\mathbb{P}(\mathbf{X}^{t_{k:k+1}})} L_{\text{1-step}}(X^{t_{k+1}}, s_\theta(X^{t_k}))\,.
\end{align*}
This imposes a stronger inductive bias that physical dynamics are Markovian, and should operate the same at any time during a trajectory.

In fact, we note that optimizing for whole trajectories may not actually not be ideal, as it can allow the simulator to learn biases which may not be hold generally. In particular, an $L$ which considers the whole trajectory means $\theta^*$ does not necessarily equal the $\theta^*_{\text{1-step}}$ that would optimize $L_{\text{1-step}}$.
This is because optimizing a capacity-limited simulator model for whole trajectories might benefit from producing greater one-step errors at certain times, in order to allow for better overall performance in the long term. For example, imagine simulating an undamped pendulum system, where the initial velocity of the bob is always zero. The physics dictate that in the future, whenever the bob returns to its initial position, it must always have zero velocity. If $\simulator_\theta$ cannot learn to approximate this system exactly, and makes mistakes on intermediate timesteps, this means that when the bob returns to its initial position it might not have zero velocity. Such errors could accumulate over time, and causes large loss under an $L$ which considers whole trajectories. The training process could overcome this by selecting $\theta^*$ which, for example, subtly encodes the initial position in the small decimal places of its predictions, which the simulator could then exploit by snapping the bob back to zero velocity when it reaches that initial position. The resulting $\simulator_{\theta^*}$ may be more accurate over long trajectories, but not generalize as well to situations where the initial velocity is not zero. This corresponds to using the predictions, in part, as a sort of memory buffer, analogous to a recurrent neural network.

Of course, a simulator with a memory mechanism can potentially offer advantages,  such as being better able to recognize and respect certain symmetries, e.g., conservation of energy and momentum. An interesting area for future work is exploring different approaches for training learnable simulators, and allowing them to store information over rollout timesteps, especially as a function for how the predictions will be used, which may favor different trade-offs between accuracy over time, what aspects of the predictions are most important to get right, generalization, etc.

\section{Supplementary Experimental Methods}
\label{supp:sec:experiments}

\subsection{Physical Domains}

\begin{center}
\begin{tabular}[p]{|l||c|c|c|c|c|c|c|c|}
	\hline
	\textbf{Domain} &  
	\multiline{15mm}{\centering\textbf{Simulator \\ (Dim.)}} & 
	\multiline{13mm}{\centering\textbf{Max. \# particles \\ (approx)}} &
	\multiline{16mm}{\centering\textbf{Trajectory \\ length}} &
	\multiline[20mm]{25mm}{\centering\textbf{\# trajectories \\ (Train/\\Validation/\\Test)}}  &
	\multiline{7mm}{\centering$\mathbf{\Delta t}$ \textbf{[ms]}} &
	\multiline{19mm}{\centering\textbf{Connectivity \\ radius}} & 
	\multiline{12mm}{\centering\textbf{Max. \# edges\\(approx)}} \\\hline\hline
    \domain{Water-3D} & SPH (3D) & 13k & 800 & 1000/100/100 & 5 & 0.035 & 230k \\\hline
    \domain{Sand-3D} & MPM (3D) & 20k & 350 & 1000/100/100 & 2.5 & 0.025 & 320k \\\hline
    \domain{Goop-3D} & MPM (3D) & 14k & 300 & 1000/100/100 & 2.5 & 0.025 & 230k \\\hline
    \domain{Water-3D-S} & SPH (3D) & 5.8k & 800 & 1000/100/100 & 5 & 0.045 & 100k \\\hline
    \domain{BoxBath} & PBD (3D) & 1k & 150 & 2700/150/150 & 16.7 & 0.08 & 17k \\\hline
    \domain{Water} & MPM (2D) & 1.9k & 1000 & 1000/30/30 & 2.5 & 0.015 & 27k \\\hline
    \domain{Sand} & MPM (2D) & 2k & 320 & 1000/30/30 & 2.5 & 0.015 & 21k \\\hline
    \domain{Goop} & MPM (2D) & 1.9k & 400 & 1000/30/30 & 2.5 & 0.015 & 19k \\\hline
    \domain{MultiMaterial} & MPM (2D) & 2k & 1000 & 1000/100/100 & 2.5 & 0.015 & 25k \\\hline
    \domain{FluidShake} & MPM (2D) & 1.3k & 2000 & 1000/100/100 & 2.5 & 0.015 & 20k \\\hline
    \domain{FluidShake-Box} & MPM (2D) & 1.5k & 1500 & 1000/100/100 & 2.5 & 0.015 & 19k \\\hline
    \domain{WaterDrop} & MPM (2D) & 1k & 1000 & 1000/30/30 & 2.5 & 0.015 & 12k \\\hline
    \domain{WaterDrop-XL} & MPM (2D) & 7.1k & 1000 & 1000/100/100 & 2.5 & 0.01 & 210k \\\hline
    \domain{WaterRamps} & MPM (2D) & 2.3k & 600 & 1000/100/100 & 2.5 & 0.015 & 26k \\\hline
    \domain{SandRamps} & MPM (2D) & 3.3k & 400 & 1000/100/100 & 2.5 & 0.015 & 32k \\\hline
    \domain{RandomFloor} & MPM (2D) & 3.4k & 600 & 1000/100/100 & 2.5 & 0.015 & 44k \\\hline
    \domain{Continuous} & MPM (2D) & 4.3k & 400 & 1000/100/100 & 2.5 & 0.015 & 47k \\\hline
\end{tabular}
\end{center}

\subsection{Implementation Details}

\xhdr{Input and output representations}
We define the input ``velocity'' as average velocity between the current and previous timestep, which is calculated from the difference in position, ${\dot{\mathbf{p}}^{t_k}\equiv \mathbf{p}^{t_{k}} - \mathbf{p}^{t_{k-1}}}$ (omitting constant $\Delta t$ for simplicity). Similarly, we define ``acceleration'' as average acceleration between the next and current timestep, ${\ddot{\mathbf{p}}^{t_k}\equiv\dot{\mathbf{p}}^{t_{k+1}} - \dot{\mathbf{p}}^{t_{k}}}$. Accelerations are thus calculated as, ${\ddot{\mathbf{p}}^{t_k} = \mathbf{p}^{t_{k+1}} - 2\mathbf{p}^{t_{k}} +\mathbf{p}^{t_{k-1}}}$.

We express the material type (water, sand, goop, rigid, boundary particle) as a particle feature, $\mathbf{a}_i$, represented with a learned embedding vector of size 16. For datasets with fixed flat orthogonal walls, instead of adding boundary particles, we add a feature to each node indicating the vector distance to each wall. Crucially, to maintain spatial translation invariance, we clip this distance to the connectivity radius $R$, achieving a similar effect to that of the boundary particles.
In \domain{FluidShake}, particle positions were provided in the coordinate frame of the container, and the container position, velocity and acceleration were provided as 6 global features.
In \domain{Continuous} a single global scalar was used to indicate the friction angle of the material.

\xhdr{Building the graph}
We construct the graph by, for each particle, finding all neighboring particles within the connectivity radius. We use a standard $k$-$d$ tree algorithm for this search. The connectivity radius was chosen, such that the number of neighbors in roughly in the range of $10-20$. We however did not find it necessary to fine-tune this parameter: All 2D scenes of the same resolution share $R=0.015$, only the high-res 2D and 3D scenes, which had substantially different particle densities, required choosing a different radius. Note that for these datasets, the radius was simply chosen once based on particle neighborhood size adn total number of edges, and was not fine-tuned as a hyperparameter.

\xhdr{Neural network parametrizations}
We also trained models where we replaced the deep encoder and decoder MLPs by simple linear layers without activations, and observed similar performance.

\subsection{Training}

\xhdr{Noise injection} Because our models take as input a sequence of states (positions and velocities), we draw independent samples $\sim \mathcal{N}(0, \sigma_v=0.0003)$, for each input state, particle and spatial dimension, before each training step. We accumulate them across time as a random walk, and use this to perturb the stack of input velocities. Based on the updated velocities, we then adjust the position features, such that $\dot{\mathbf{p}}^{t_k} \equiv \mathbf{p}^{t_{k}} - \mathbf{p}^{t_{k-1}}$ is maintained, for consistency. We also experimented with other types of noise accumulation, as detailed in Section~\ref{supp:sec:results}.

Another way to address differences in training and test input distributions is to, during training, provide the model with its own predictions by rolling out short sequences. \citet{ummenhofer2020lagrangian}, for example, train with two-step predictions. However computing additional model predictions are more expensive, and in our experience may not generalize to longer sequences as well as noise injection.

\xhdr{Normalization} To normalize our inputs and targets, we compute the dataset statistics during training. Instead of using moving averages, which could shift in cycles during training, we instead build exact mean and variance for all of the input and target particle features up seen up to the current training step $l$, by accumulating the sum, the sum of the squares and the total particle count. The statistics are computed after noise is applied to the inputs. 

\xhdr{Loss function and optimization procedures} 
We load the training trajectories sequentially, and use them to generate input and target pairs (from a 1000-step long trajectory we generate 995 pairs, as we condition on 5 past states), and sample input-target pairs from a shuffle buffer of size 10k. 
The acceleration targets are computed from the sequence of positions before adding noise to the input, and then adjusted by removing the input velocity noise accumulated at the final step of the random walk. By doing so, the model learns to predict an output acceleration that, after integrating with the Euler integrator ($\Delta t=1$), yields a next-step velocity that matches the next-step ground-truth velocity, hence correcting for the noise in input velocity. This is equivalent to defining the loss on next-step ground-truth velocity.
Rigid obstacles, such as the ramps in \domain{Water-Ramps}, are represented as \emph{boundary particles}. Those particles are treated identical to regular particles, but they are masked out of the loss.

Due to normalization of predictions and targets, our prediction loss is normalized, too. This allows us to choose a scale-free learning rate, across all datasets. 
To optimize the loss, we use the Adam optimizer~\cite{kingma2014adam} (a form of stochastic gradient descent) with a nominal mini-batch size of 2 examples, averaging the loss for all particles in the batch. We performed a maximum of 20M gradient update steps, with an exponentially decaying learning rate, $\alpha(j)$, where on the $j$-th gradient update step, ${\alpha(j) = \alpha_{\mathrm{final}} + (\alpha_{\mathrm{start}} - \alpha_{\mathrm{final}}) \cdot 0.1^{(j \cdot 5 \cdot 10^6)}}$, with $\alpha_{\mathrm{start}} = 10^{-4}$ and $\alpha_{\mathrm{final}} = 10^{-6}$. While models can train in significantly less steps, we avoid aggressive learning rates to reduce variance across datasets and make comparisons across settings more fair.

We train our models using second generation TPUs and V100 GPUs interchangeably. For our datasets, we found that training time per example with a single TPU core or a single V100 GPU was about the same. TPUs allowed for faster training through fast batch parallelism (each of the two training examples in the batch runs on a separate TPU core in the same TPU chip). Furthermore, since TPU cores require fixed size tensors, instead of just padding each training example up to a maximum number of nodes/edges, we set the fixed size to correspond to the largest graph in the dataset, and, at each step, build a larger minibatch using multiple training examples whenever they would fit within the set fixed size, before adding the padding. This yielded an effective batch size between 1 (large examples) and 3 examples (small examples) per device (2 to 6 examples per batch when batch parallelism is taken into account) and is equivalent to setting a mini batch size in terms of total number of particles per batch. For easier comparison, we also replicated this procedure on the GPU training. 

Additionally, for our largest systems (> 100k edges) we also used model parallelism (a single training example distributed over multiple TPU cores)\cite{kumar2019scale}, over a total of 16 TPU cores per example (16 TPU chips in total, given the batch parallelism of 2 and that each TPU chip has two TPU cores).

\subsection{Distributional Evaluation Metrics}
An MSE metric of zero indicates that the model perfectly predicts where each particle has traveled. However, if the model's predicted positions for particles A and B exactly match true positions of particles B and A, respectively, the MSE could be high, even though the predicted and true distributions of particles match. So we also explored two metrics that are invariant under particle permutations, by measuring differences between the distributions of particles: optimal transport (OT)~\cite{villani2003topics} using 2D or 3D Wasserstein distance and approximated by the Sinkhorn Algorithm~\cite{cuturi2013sinkhorn}, and maximum mean discrepancy (MMD)~\cite{gretton2012kernel} with a Gaussian kernel bandwidth of ${\sigma = 0.1}$. These distributional metrics may be more appropriate when the goal is to predict what regions of the space will be occupied by the simulated material, or when the particles are sampled from some continuous representation of the state and there are no ``true'' particles to compare predictions to. We will analyze some of our results using those metrics in Section~\ref{supp:sec:results}.

\section{Supplementary Results}
\label{supp:sec:results}

\subsection{Architectural Choices with Minor Impact on Performance} 

\begin{figure}
  \centering
  \includegraphics[trim=0 0 0 0, clip, width=.70\textwidth]{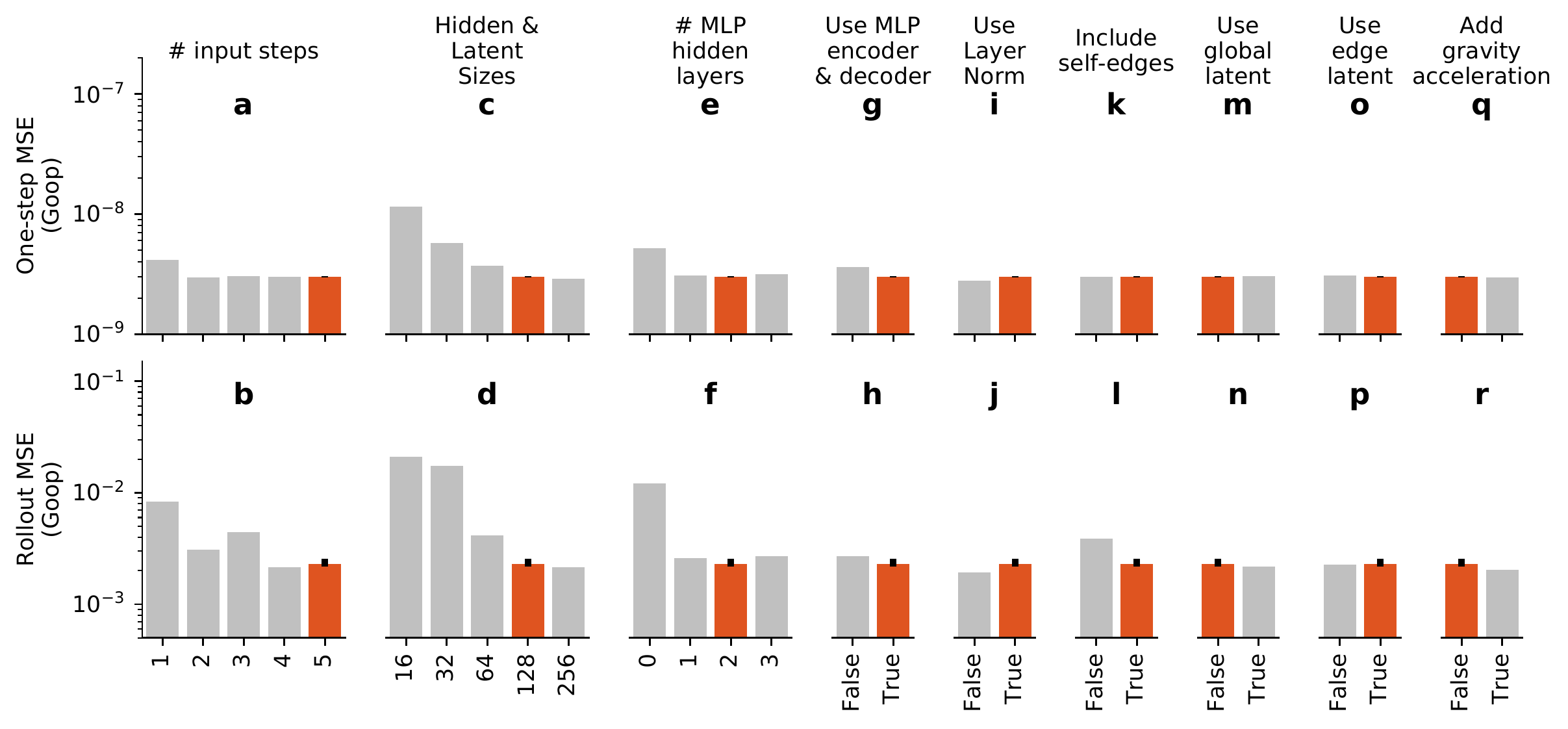}
  \caption{Additional ablations (grey) on the \domain{Goop} dataset compared to our default model (red). Error bars display lower and higher quartiles, and are shown for the default parameters. The same vertical limits from \figref{fig:main_ablations} are reused for easier qualitative scale comparison.}
  \label{supp:fig:other_ablations}
\end{figure}

We provided additional ablation scans on the \domain{Goop} dataset in \figref{supp:fig:other_ablations}, which show that the model is robust to typical hyperparameter changes.

\xhdr{Number of input steps $C$}
(\figref{supp:fig:noise_ablations}a,b) We observe a significant improvement from conditioning the model on just the previous velocity ($C=1$) to the two most recent velocities ($C=2$), but find similar performance for larger values of $C$. All our models use $C=5$. Note that because we approximate the velocity as the finite difference of the position, the model requires the most recent $C+1$ positions to compute the last $C$ velocities.

\xhdr{Latent and MLP layer sizes}
(\figref{supp:fig:noise_ablations}c,d) The performance does not change much as function of the latent and MLP hidden sizes, for sizes of 64 or larger.

\xhdr{Number of MLP hidden layers}
(\figref{supp:fig:noise_ablations}e,f) Except for the case of zero hidden layers (linear layer), the performance does not change much as a function of the MLP depth.

\xhdr{Use MLP encoder \& decoder}
(\figref{supp:fig:noise_ablations}g,h) We replaced the MLPs used in the encoder and decoder by linear layers (single matrix multiplication followed by a bias), and observed no significant changes in performance.

\xhdr{Use LayerNorm}
(\figref{supp:fig:noise_ablations}i,j) In small datasets, we typically observe slightly better performance when LayerNorm~\cite{ba2016layer} is not used, however, enabling it provides additional training stability for the larger datasets.

\xhdr{Include self-edges}
(\figref{supp:fig:noise_ablations}k,l) The model performs similarly regardless of whether self-edges are included in the message passing process or not.

\xhdr{Use global latent}
(\figref{supp:fig:noise_ablations}m,n) We enable the global mechanisms of the GNs. This includes both, explicit input global features (instead of appending them to the nodes) and a global latent state that updates after every message passing step using aggregated information from all edges and nodes. We do not find significant differences when doing this, however we speculate that generalization performance for systems with more particles than used during training would be affected.

\xhdr{Use edges latent}
(\figref{supp:fig:noise_ablations}o,p) We disabled the updates to the latent state of the edges that is performed at each edge message passing iteration, but found no significant differences in performance.

\xhdr{Add gravity acceleration}
(\figref{supp:fig:noise_ablations}q,r) We attempted adding gravity accelerations to the model outputs in the update procedure, so the model would not need to learn to predict a bias acceleration due to gravity, but found no significant performance differences.

\subsection{Noise-Related Training Parameters}

\begin{figure}
  \centering
  \includegraphics[trim=0 0 0 0, clip, width=.4\textwidth]{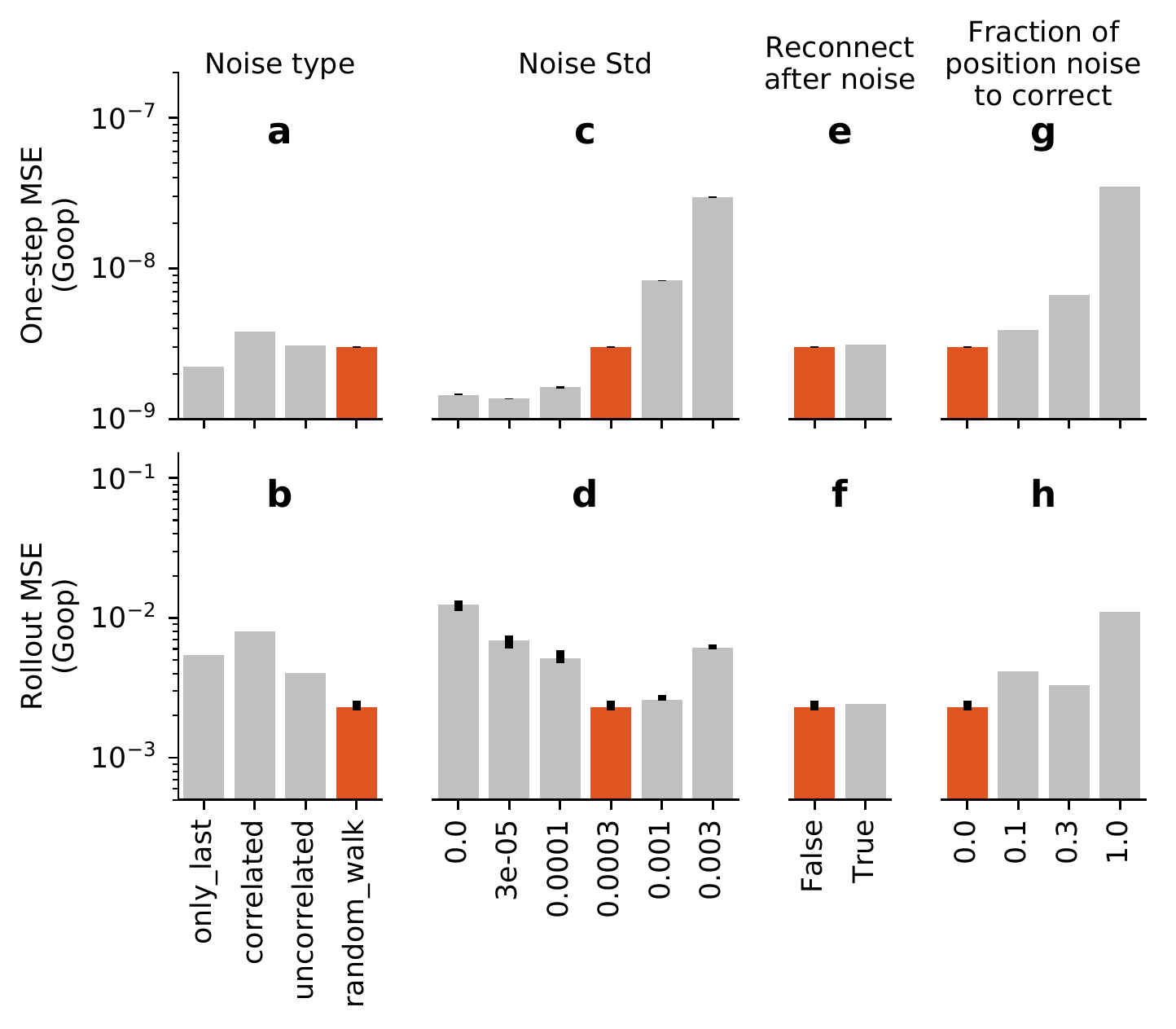}
  \caption{Noise-related training variations (grey) on the \domain{Goop} dataset compared to our default model (red). Error bars display lower and higher quartiles, and are shown for the default parameters. The same vertical limits from \figref{fig:main_ablations} are reused for easier qualitative scale comparison.}
  \label{supp:fig:noise_ablations}
\end{figure}

We provide some variations related to how we add noise to the input data on the \domain{Goop} dataset in \figref{supp:fig:noise_ablations}.

\xhdr{Noise type}
(\figref{supp:fig:noise_ablations}a,e) We experimented with 4 different modes for adding noise to the inputs. \emph{only\_last} adds noise only to the velocity of the most recent state in the input sequence of states. \emph{correlated} draws a single per-particle and per-dimension set of noise samples, and applies the same noise to the velocities of all input states in the sequence. \emph{uncorrelated} draws independent noise for the velocity of each input state. \emph{random\_walk} draws noise for each input state, adding it to the noise of the previous state in sequence as in a random random walk, as an attempt to simulate accumulation of error in a rollout. In all cases the input states positions are adjusted to maintain $\dot{\mathbf{p}}^{t_k} \equiv \mathbf{p}^{t_{k}} - \mathbf{p}^{t_{k-1}}$. To facilitate the comparison, the variance of the generated noise is adjusted so the variance of the velocity noise at the last step is constant. We found the best rollout performance for \emph{random\_walk} noise type.

\xhdr{Noise Std}
(\figref{supp:fig:noise_ablations}b,f) This is described in the main text, included here for completeness.

\xhdr{Reconnect graph after noise}
(\figref{supp:fig:noise_ablations}c,g) We found that the performance did not change regardless of whether we recalculated the connectivity of the graph after applying noise to the positions or not.

\xhdr{Fraction of position noise to correct}
(\figref{supp:fig:noise_ablations}d,h) In the process of corrupting the input position and velocity features with noise, we adjust the target accelerations such as the acceleration predicted by the model would compensate for the noise in input velocity  (0\%). For this ablation we modify the target accelerations such as the model learns to predict accelerations that would instead correct the noise in the input position (100\%). Note that this happens implicitly when the loss is defined directly on next-step ground-truth position, regardless of whether the inputs are perturbed with noise \cite{sanchez2018graph} or the inputs have noise due to model error after a rollout \cite{ummenhofer2020lagrangian}. We also investigate two intermediary points (10\% and 30\%), which implies training the model to correct a mix of the position and the velocity noise. Note that since the model uses a simple Euler update to compute output position from output velocity, it is impossible to exactly correct for both position and velocity noise; we have to choose one or the other (or a compromise in between). \figref{supp:fig:noise_ablations}d,h show that asking the model to correct for the position noise instead of the velocity noise leads to worse performance.

\begin{figure}
  \centering
  \includegraphics[trim=0 0 0 0, clip, width=.7\textwidth]{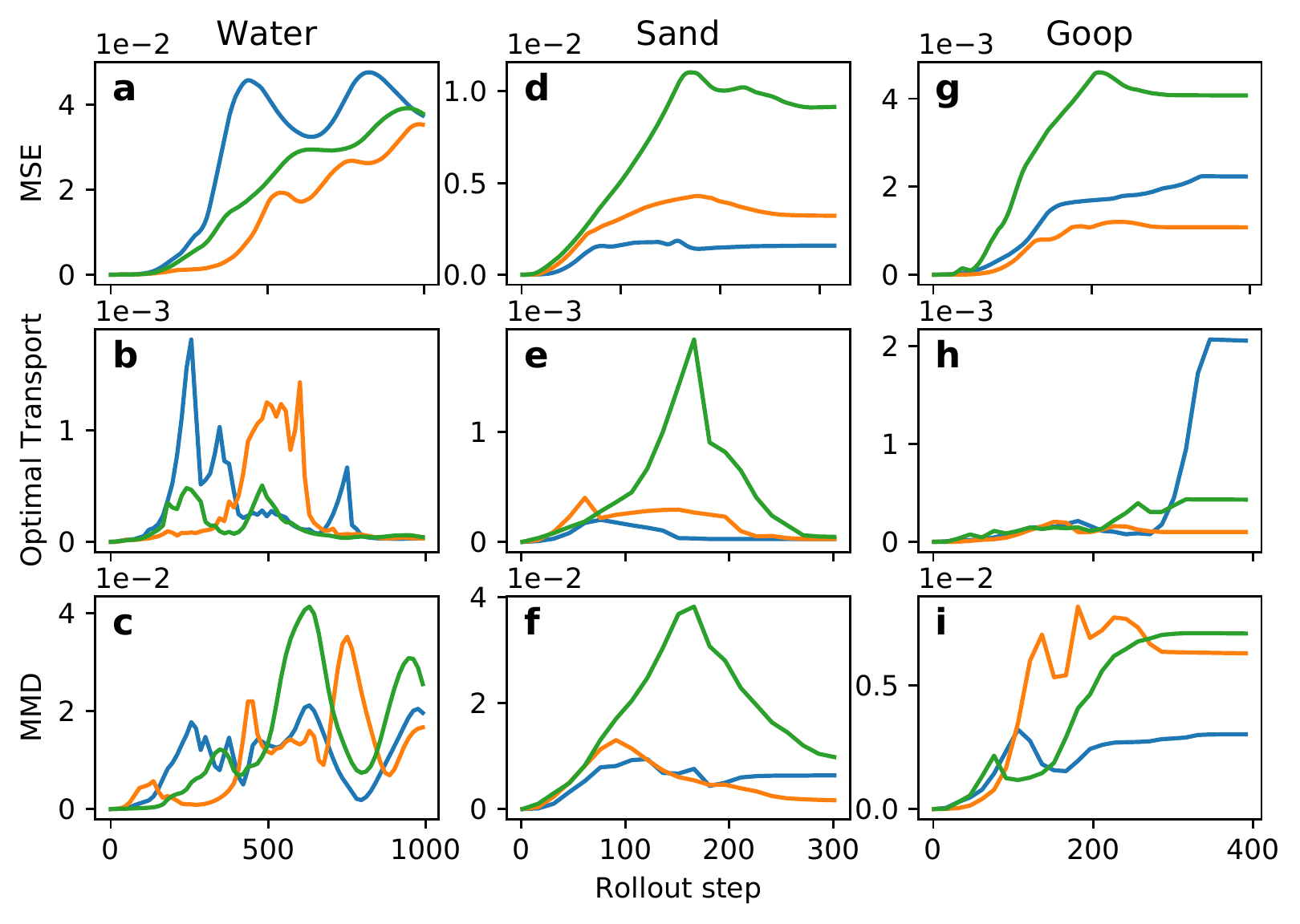}
  \caption{Rollout error of one of our models as a function of time for water, sand and goop, using MSE, Optimal transport and MMD as metrics. Figures show curves for 3 trajectories in the evaluation datasets (colored curves). MSE tends to grow as function of time due to the chaotic character of the systems. Optimal transport and MMD, which are invariant to particle permutations, tend to decrease for \domain{Water} and \domain{Sand} towards the end of the rollout as they equilibrate towards a single consistent group of particles, due to the lower friction/viscosity. For \domain{Goop}, which can remain in separate clusters in a chaotic manner, the Optimal Transport error can still be very high.}
  \label{supp:fig:metrics_with_time}
\end{figure}

\subsection{Distributional Evaluation Metrics}

\begin{figure}
  \centering
  \includegraphics[trim=0 0 0 0,clip,    width=.6\textwidth]{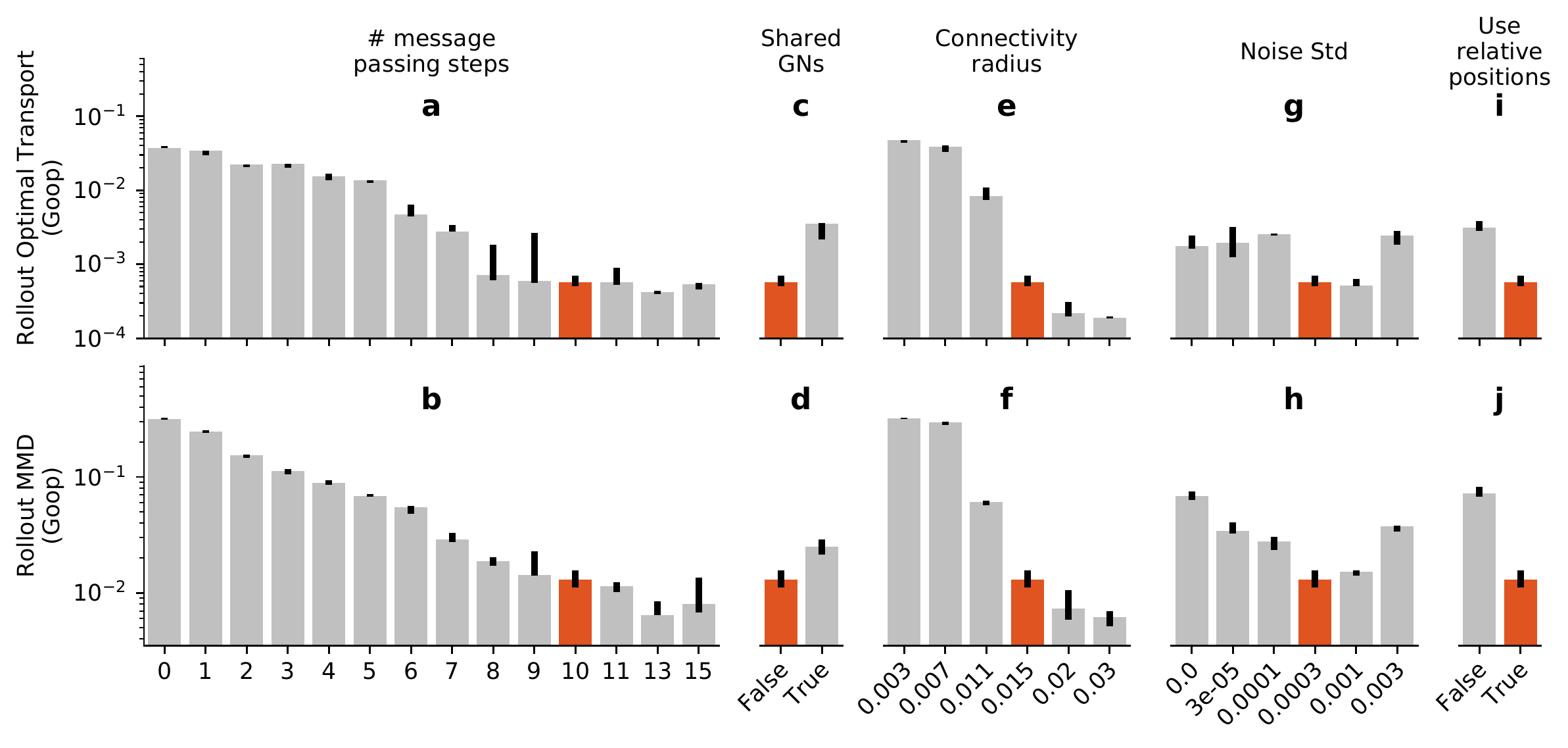}
  \caption{Effect of different ablations on Optimal Transport (top) and Maximum Mean Discrepancy (bottom) rollout errors. Bars show the median seed performance averaged across the entire test dataset. Error bars display lower and higher quartiles, and are shown for the default parameters.}
  \label{fig:main_ablations_distributional}
\end{figure}

Generally we find that MSE and the distributional metrics lead to generally similar conclusions in our analyses~(see~\figref{supp:fig:metrics_with_time}), though we notice that differences in the distributional metrics' values for qualitatively ``good'' and ``bad'' rollouts can be more prominent, and match more closely with our subjective visual impressions. \figref{fig:main_ablations_distributional} shows the rollout errors as a function of the key architectural choices from \figref{fig:main_ablations} using these distributional metrics.

\subsection{Quantitative Results on all Datasets}

\begin{center}
\begin{tabular}[p]{|l||c|c|c|c|c|c|}
	\hline
	&
	\multicolumn{2}{c|}{\textbf{Mean Squared Error}} &
	\multicolumn{2}{c|}{\textbf{Optimal Transport}} &
	\multicolumn{2}{c|}{\multiline[20mm]{25mm}{\centering\textbf{Maximum Mean \\ Discrepancy \\ ($\sigma=0.1$)}}}\\
	\textbf{Domain} &  
	\multiline[15mm]{15mm}{\centering\textbf{One-step} \\ $\times10^{-9}$}  & \multiline[15mm]{15mm}{\centering\textbf{Rollout} \\ $\times10^{-3}$} &
	\multiline[15mm]{15mm}{\centering\textbf{One-step} \\ $\times10^{-9}$}  & \multiline[15mm]{15mm}{\centering\textbf{Rollout} \\ $\times10^{-3}$} &
	\multiline[15mm]{15mm}{\centering\textbf{One-step} \\ $\times10^{-9}$} &
	\multiline[15mm]{15mm}{\centering\textbf{Rollout} \\ $\times10^{-3}$}
	\\\hline\hline
    \domain{Water-3D} &  8.66 & 10.1 & 26.5 & 0.165 & 7.32 & 0.368 \\\hline
    \domain{Sand-3D} &  1.42 & 0.554 & 4.29 & 0.138 & 11.9 & 2.67 \\\hline
    \domain{Goop-3D} &  1.32 & 0.618 & 4.05 & 0.208 & 22.4 & 5.13 \\\hline
    \domain{Water-3D-S} &  9.66 & 9.52 & 29.9 & 0.222 & 6.9 & 0.192 \\\hline
    \domain{BoxBath} &  54.5 & 4.2 & -- & -- & -- & -- \\\hline
    \domain{Water} &  2.82 & 17.4 & 6.19 & 0.468 & 10.6 & 7.66 \\\hline
    \domain{Sand} &  6.23 & 2.37 & 11.8 & 0.193 & 32.6 & 6.79 \\\hline
    \domain{Goop} &  2.91 & 1.89 & 6.14 & 0.419 & 20.3 & 7.76 \\\hline
    \domain{MultiMaterial} &  1.81 & 16.9 & -- & -- & -- & -- \\\hline
    \domain{FluidShake} &  2.1 & 20.1 & 4.13 & 0.591 & 12.1 & 9.84 \\\hline
    \domain{FluidShake-Box} &  1.33 & 4.86 & -- & -- & -- & -- \\\hline
    \domain{WaterDrop} &  1.52 & 7.01 & 3.31 & 0.273 & 11.1 & 5.99 \\\hline
    \domain{WaterDrop-XL} &  1.23 & 14.9 & 3.03 & 0.209 & 11.6 & 4.34 \\\hline
    \domain{WaterRamps} &  4.91 & 11.6 & 10.3 & 0.507 & 14.3 & 7.68 \\\hline
    \domain{SandRamps} &  2.77 & 2.07 & 6.13 & 0.187 & 15.7 & 4.81 \\\hline
    \domain{RandomFloor} &  2.77 & 6.72 & 5.8 & 0.276 & 14 & 3.53 \\\hline
    \domain{Continuous} &  2.06 & 1.06 & 4.63 & 0.0709 & 23.1 & 3.57 \\\hline
\end{tabular}
\end{center}

\subsection{Quantitative Generalization Results on \domain{Continuous} Domain} 

\begin{figure}[h]
  \centering
  \includegraphics[trim=0 0 0 0,clip,    width=.6\textwidth]{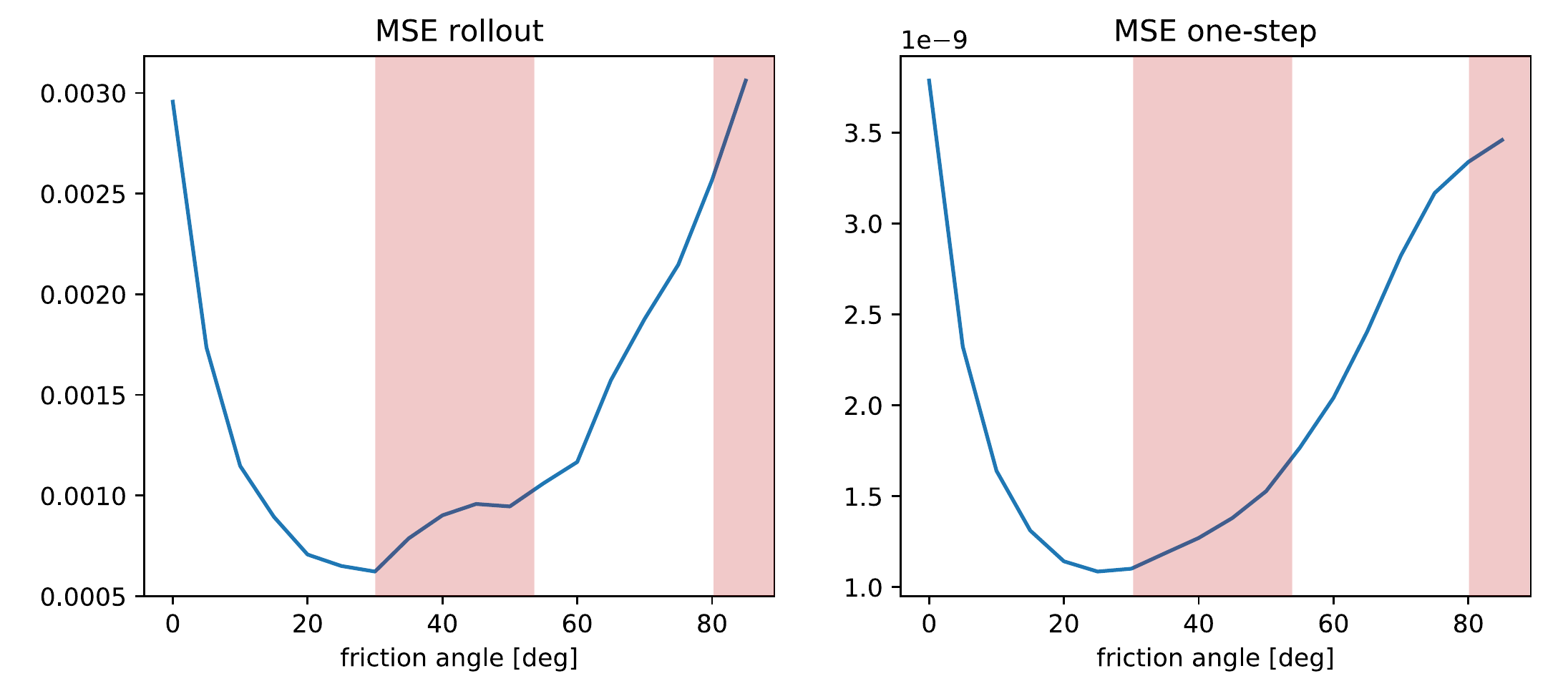}
  \caption{Rollout and one-step error as a function of the friction angle in the \domain{Continuous} domain. Regions highlighted in red correspond to values of the friction angle not observed during training. Our results show that a model trained in the $[0^{\circ}, 30^{\circ}]$ and $[55^{\circ}, 80^{\circ}]$ ranges can produce good predictions across all friction angles (see also \vidcont{videos}), with only marginally higher errors in the $[30^{\circ}, 55^{\circ}]$ range that was not seen during training. Note that the dynamics at very low and very high friction angles are simply more complicated and harder to learn, hence the higher error.}
  \label{fig:continuous_generalization}
\end{figure}

See \figref{fig:continuous_generalization}.

\subsection{Inference Times}

The following table compares the performance of our learned GNS model (evaluated on a single V100 GPU) to the performance of the simulator used to generate the data (run on a 6-core workstation CPU) for each of the datasets. The learned GNS model has inference times comparable to that of the ground truth simulator used to generate the data. Note that these results are purely informative and calculated post-hoc, as the main goal of this work was not to improve simulation time. We expect better performance could be achieved by making predictions for longer time steps, or using optimized implementations for neighborhood graph calculation (which we evaluated out-of-graph on the CPU using KDTrees).
\begin{center}
\begin{tabular}[p]{|l||c|c|c|c|c|}
	\hline
	\textbf{Domain} &  
	\multiline{15mm}{\centering\textbf{Simulator \\ (Dim.)}} & 
	\multiline{15mm}{\centering\textbf{Mean \# particles\\per graph\\(approx)}} &
	\multiline{15mm}{\centering\textbf{Mean \# edges\\per graph\\(approx)}} &
	\multiline{15mm}{\centering\textbf{Simulator time per step [s]}} &
	\multiline[18mm]{56mm}{\centering\textbf{Learned GNS time per step \\ including neighborhood computation \\ {[s]} (relative to simulator)}} \\\hline\hline
    \domain{Water-3D} & SPH (3D) & 7.8k & 110k & 0.104 & 0.358 (345\%) \\\hline
    \domain{Sand-3D} & MPM (3D) & 9.8k & 140k & 0.221 & 0.336 (152\%) \\\hline
    \domain{Goop-3D} & MPM (3D) & 7.8k & 120k & 0.199 & 0.247 (124\%) \\\hline
    \domain{Water-3D-S} & SPH (3D) & 3.8k & 55k & 0.053 & 0.0683 (129\%) \\\hline
    \domain{BoxBath} & PBD (3D) & 1k & 15k & -- & 0.0475 (--) \\\hline
    \domain{Water} & MPM (2D) & 1.1k & 12k & 0.037 & 0.0579 (156\%) \\\hline
    \domain{Sand} & MPM (2D) & 1.2k & 11k & 0.045 & 0.048 (107\%) \\\hline
    \domain{Goop} & MPM (2D) & 1k & 9.2k & 0.04 & 0.0301 (75.1\%) \\\hline
    \domain{MultiMaterial} & MPM (2D) & 1.6k & 16k & 0.049 & 0.0595 (121\%) \\\hline
    \domain{FluidShake} & MPM (2D) & 1.3k & 13k & 0.039 & 0.0257 (65.8\%) \\\hline
    \domain{FluidShake-Box} & MPM (2D) & 1.4k & 13k & 0.048 & 0.133 (277\%) \\\hline
    \domain{WaterDrop} & MPM (2D) & 0.6k & 4.8k & 0.05 & 0.0256 (51.3\%) \\\hline
    \domain{WaterDrop-XL} & MPM (2D) & 4.3k & 83k & 0.166 & 0.192 (116\%) \\\hline
    \domain{WaterRamps} & MPM (2D) & 1.5k & 13k & 0.071 & 0.0506 (71.3\%) \\\hline
    \domain{SandRamps} & MPM (2D) & 2.3k & 21k & 0.077 & 0.0691 (89.8\%) \\\hline
    \domain{RandomFloor} & MPM (2D) & 2.3k & 24k & 0.076 & 0.0634 (83.4\%) \\\hline
    \domain{Continuous} & MPM (2D) & 2.4k & 22k & 0.072 & 0.0919 (128\%) \\\hline
\end{tabular}
\end{center}

The table below shows the inference time for batches padded to a maximum size with pre-computed neighborhoods (which is equivalent to inference on a single large graph). Comparing to the previous table, we indeed observe that most of the computation time was spent on neighborhood computation rather than on graph neural network inference.

\begin{center}
\begin{tabular}[p]{|l||c|c|c|}
	\hline
	\textbf{Domain} &  
	\multiline{16mm}{\centering\textbf{\# particles in batch }} &
	\multiline{16mm}{\centering\textbf{\# edges in batch }} &
	\multiline[15mm]{80mm}{\centering\textbf{Learned GNS time per step \\ without neighborhood computation [s] \\ (relative to learned GNS in previous table)\footnotemark}}\\\hline\hline
    \domain{Water-3D} & 14k & 245k & 0.071 (19.8\%) \\\hline
    \domain{Sand-3D} & 19k & 320k & 0.086 (25.6\%) \\\hline
    \domain{Goop-3D} & 15k & 230k & 0.109 (44.2\%) \\\hline
    \domain{Water-3D-S} & 6k & 120k & 0.04 (58.6\%) \\\hline
    \domain{BoxBath} & 1k & 18k & 0.017 (35.8\%) \\\hline
    \domain{Water} & 2k & 31k & 0.025 (43.2\%) \\\hline
    \domain{Sand} & 2k & 21k & 0.018 (37.5\%) \\\hline
    \domain{Goop} & 2k & 21k & 0.019 (63.2\%) \\\hline
    \domain{MultiMaterial} & 2k & 27k & 0.018 (30.3\%) \\\hline
    \domain{FluidShake} & 1.4k & 23k & 0.017 (66.2\%) \\\hline
    \domain{FluidShake-Box} & 1.5k & 20k & 0.019 (14.3\%) \\\hline
    \domain{WaterDrop} & 2k & 18k & 0.023 (89.7\%) \\\hline
    \domain{WaterDrop-XL} & 8k & 300k & 0.057 (29.7\%) \\\hline
    \domain{WaterRamps} & 2.5k & 28k & 0.017 (33.6\%) \\\hline
    \domain{SandRamps} & 3.5k & 35k & 0.023 (33.3\%) \\\hline
    \domain{RandomFloor} & 3.5k & 46k & 0.023 (36.3\%) \\\hline
    \domain{Continuous} & 5k & 50k & 0.033 (35.9\%) \\\hline
\end{tabular}
\end{center}
\footnotetext{Note that these batches are equivalent to running a single graph larger than those in the previous table, so ``(relative to learned GNS in previous table)'' only provides an upper bound on the fraction of time spent on graph neural network inference.}

\subsection{Example Failure Cases}
In \vidfail{this video}, we show two of the failure cases we sometimes observe with the GNS model. In the \domain{BoxBath} domain we found that our model could accurately predict the motion of a rigid block, and maintain its shape, without requiring explicit mechanisms to enforce solidity constraints or providing the rest shape to the network. However, we did observe limits to this capability in a harder version of \domain{BoxBath}, which we called \domain{FluidShake-Box}, where the container is vigorously shaken side to side, over a rollout of 1500 timesteps. Towards the end of the trajectory, we observe that the solid block starts to deform. 
We speculate the reason for this is that GNS has to keep track of the block's original shape, which can be difficult to achieve over long trajectories given an input of only 5 initial frames.

In the second example, a \emph{bad} seed of our model trained on the \domain{Goop} domain predicts a blob of goop stuck to the wall instead of falling down. We note that in the training data, the blobs do sometimes stick to the wall, though it tends to be closer to the floor and with different velocities. We speculate that the intricacies of static friction and adhesion may be hard to learn---to learn this behaviour more robustly, the model may need more exposure to fall versus sticking phenomena.

\section{Supplementary Baseline Comparisons}
\label{supp:sec:baselines}

\subsection{Continuous Convolution (CConv)}
Recently~\citet{ummenhofer2020lagrangian} presented Continuous Convolution~(CConv) as a method for particle-based fluid simulation. We show that CConv can also be understood in our framework, and compare CConv to our approach on several tasks.

\xhdr{Interpretation}

\nopagebreak
While \citet{ummenhofer2020lagrangian} state that ``Unlike previous approaches, we do not build an explicit graph structure to connect the particles but use spatial convolutions as the main differentiable operation that relates particles to their neighbors.'', we find we can express CConv  (which itself is a generalization of CNNs) as a GN~\cite{battaglia2018relational} with a specific type of edge update function.

CConv relates to CNNs (with stride of 1) and GNs in two ways. First, in CNNs, CConv, and GNs, each element (e.g., pixel, feature vector, particle) is updated as a function of its neighbors. In CNNs the neighborhood is fixed and defined by the kernel's dimensions, while in CConv and GNs the neighborhood varies and is defined by connected edges (in CConv the edges connect to nearest neighbors).

Second, CNNs, CConv, and GNs all apply a function to element $i$'s neighbors, $j \in \mathcal{N}(i)$, pool the results from within the neighborhood, and update element $i$'s representation. In a CNN, this is computed as,~$f'_i = \sigma\left(\mathbf{b} + \sum_{j \in \mathcal{N}(i)} W(\bm{\tau}_{i,j}) f_j \right)$, where $W(\bm{\tau}_{i,j})$ is a matrix whose parameters depend on the displacement between the grid coordinates of $i$ and $j$, $\bm{\tau}_{i,j} = \mathbf{x}_j - \mathbf{x}_i$ (and $\mathbf{b}$ is a bias vector, and $\sigma$ is a non-linear activation). Because there are a finite set of~$\bm{\tau}_{i,j}$ values, one for each coordinate in the kernel's grid, there are a finite set of~$W(\bm{\tau}_{i,j})$ parameterizations.

CConv uses a similar formula, except the particles' continuous coordinates mean a choice must be made about how to parameterize~$W(\bm{\tau}_{i,j})$. 
Like in CNNs, CConv uses a finite set of distinct weight matrices,~$\hat{W}(\hat{\bm{\tau}}_{i,j})$, associated with the discrete coordinates,~$\hat{\bm{\tau}}_{i,j}$, on the kernel's grid. For the continuous input~$\bm{\tau}_{i,j}$, the nearest $\hat{W}(\bm{\tau}_{i,j})$ are interpolated by the fractional component of~$\bm{\tau}_{i,j}$. In 1D this would be linear interpolation, $W(\bm{\tau}_{i,j}) = (1 - d)\, \hat{W}(\lfloor\bm{\tau}_{i,j}\rfloor) + d\, \hat{W}(\lceil\bm{\tau}_{i,j}\rceil))$, where $d = \bm{\tau}_{i,j} - \lfloor\bm{\tau}_{i,j}\rfloor$. In 3D, this is trilinear interpolation.

A GN can implement CNN and CConv computations by representing~$\bm{\tau}_{i,j}$ using edge attributes, $\mathbf{e}_{i,j}$, and an edge update function which uses independent parameters for each $\bm{\tau}_{i,j}$, i.e., $\mathbf{e}'_{i,j} = \phi^e(\mathbf{e}_{i,j}, \mathbf{v}_i, \mathbf{v}_j) = \phi^e_{\bm{\tau}_{i,j}}(\mathbf{v}_j)$. 
Beyond their displacement-specific edge update function, CNNs and CConv are very similar to how graph convolutional networks (GCN)~\cite{kipf2016semi} work.
The full CConv update as described in~\citet{ummenhofer2020lagrangian} is,
$f'_i = \frac{1}{\psi\left(\mathbf{x}_i\right)} \sum_{j \in \mathcal{N}\left(\mathbf{x}_i, R\right)}  a\left(\mathbf{x}_j, \mathbf{x}_i\right) f_j \; g\left(\Lambda\left(\mathbf{x}_j - \mathbf{x}_i\right)\right)$.
In particular, it indexes into the weight matrices via a polar-to-Cartesian coordinate transform, $\Lambda$, to induce a more radially homogeneous parameterization. It also uses a weighted sum over the particles in a neighborhood, where the weights, $a(\mathbf{x}_j, \mathbf{x}_i)$, are proportional to the distance between particles. And it includes a normalization, $\psi(\mathbf{x}_i)$, for neighborhood size, they set it to 1.

\xhdr{Performance comparisons}
\nopagebreak

We implemented the CConv model, loss and training procedure as described by \citet{ummenhofer2020lagrangian}.
For simplicity, we only tested the CConv model on datasets with flat walls, rather than those with irregular geometry. This way we could omit the initial convolution with the boundary particles and instead give the fluid particles additional simple features indicating the vector distance to each wall, clipped by the radius of connectivity, as in our model. This has the same spatial constraints as CConv with boundary particles in the wall, and should be as or more informative than boundary particles for square containers. Also, for environments with multiple materials, we appended a particle type learned embedding to the input node features. 

To be consistent with~\citet{ummenhofer2020lagrangian}, we used their batch size of 16, learning rate decay of $10^{-3}$ to $10^{-5}$ for 50k iterations, and connectivity radius of 4.5x the particle radius. We were able to replicate their results on PBD/FLeX and SPH simulator datasets similar to the datasets presented in their paper.  
To allow a fair comparison when evaluating on our MPM datasets, we performed additional hyperparameter sweeps over connectivity particle radius, learning rate, and number of training iterations using our \domain{Goop} dataset. We used the best-fitting parameters on all datasets, analogous to how we selected hyperparameters for our GNS model.

We also implemented variations of CConv which used noise to corrupt the inputs during training (instead of using 2-step loss), as we did with GNS. We found that the noise improved CConv's rollout performance on most datasets. 
In our comparisons, we always report performance for the best-performing CConv variant.

Our qualitative results show CConv can learn to simulate sand reasonably well. But it struggled to accurately simulate solids with more complex fine grained dynamics. In the \domain{BoxBath} domain, CConv simulated the fluid well, but struggled to keep the box's shape intact. 
In the \domain{Goop} domain, CConv struggled to keep pieces of goop together and handle the rest state, while in \domain{MultiMaterial} it exhibited local ``explosions'', where regions of particles suddenly burst outward (see \vidcconv{video}). 

Generally CConv's performance is strongest for simulating water-like fluids, which is primarily what it was applied to in the original paper. However it still did not match our GNS model, and for other materials, and interactions between different materials, it was clearly not as strong. This is not particularly surprising, given that our GNS is a more general model, and our neural network implementation has higher capacity on several axes, e.g., more message-passing steps, pairwise interaction functions, more flexible function approximators (MLPs with multiple internal layers versus single linear/non-linear layers in CConv).

\subsection{DPI}
We trained our model on the \citet{li2018learning}'s \domain{BoxBath} dataset, and directly compared the qualitative behavior to the authors' demonstration video in \vidboxbath{this comparison}. To provide a fair comparison to DPI, we show a model conditioned on just the previous velocity ($C$=1) in the above comparison video\footnote{We also ran experiments with C=5 and did not find any meaningful difference in performance. The results in \tabref{fig:mse} and the corresponding example video are run with C=5 for consistency with our other experiments}.
While DPI requires a specialized hierarchical mechanism and forced all box particles to preserve their relative displacements with each other, our GNS model faithfully represents the the ground truth trajectories of both water and solid particles without any special treatment. The particles making up the box and water are simply marked as a two different materials in the input features, similar to our other experiments with sand, water and goop. We also found that our model seems to also be more accurate when predicting the fluid particles over the long rollout, and it is able to perfectly reproduce the layering effect for fluid particles at the bottom of the box that exists in the ground truth data. 

\end{document}